\documentclass[10pt,twocolumn,letterpaper]{article}

\usepackage{cvpr}

\makeatletter
\@namedef{ver@everyshi.sty}{}
\makeatother

\usepackage{graphicx}
\usepackage{amsmath}
\usepackage{amssymb}
\usepackage{booktabs}

\usepackage{tikz}
\usepackage{comment}
\usepackage{color}
\usepackage{tabularx}
\usepackage{colortbl}
\usepackage{tabu}
\usepackage{multirow}
\usepackage{setspace}
\usepackage[font=small,labelsep=period]{caption}

\makeatletter
\renewcommand{\fnum@figure}{Figure \thefigure}
\makeatother

\newcolumntype{Y}{>{\centering\arraybackslash}X}
\newcolumntype{R}{>{\raggedleft\arraybackslash}X}
\newcolumntype{L}{>{\raggedright\arraybackslash}X}

\newcommand{\mytilde}{\raise.17ex\hbox{$\scriptstyle\sim$}}

\definecolor{bblue}{rgb}{0.12,0.583,0.78}
\definecolor{ccol}{rgb}{0.91,0.91,0.91}

\definecolor{avgcol}{rgb}{1.0,0.91,0.722}
\definecolor{arcol}{rgb}{0.765,0.878,0.812}
\definecolor{timecol}{rgb}{0.941,0.749,0.737}

\usepackage{xspace}

\newcommand\customparagraph[1]{\vspace{0.7em}\noindent\textbf{#1}}

\usepackage[pagebackref=true,breaklinks=true,letterpaper=true,colorlinks=true,urlcolor=bblue,bookmarks=false,allcolors=black]{hyperref}

\usepackage[capitalize]{cleveref}
\crefname{section}{Sec.}{Secs.}
\Crefname{section}{Section}{Sections}
\Crefname{table}{Table}{Tables}
\crefname{table}{Tab.}{Tabs.}

\def\cvprPaperID{*****}
\def\confName{CVPR}
\def\confYear{2022}

\begin{document}

\title{BOP Challenge 2022 on Detection, Segmentation \\ and Pose Estimation of Specific Rigid Objects}

\newcommand{\namesep}{\hspace{1.0em}}
\author{
 Martin Sundermeyer$^{1,2}$\namesep
 Tom{\'a}{\v{s}}~Hoda{\v{n}}$^{3}$\namesep
 Yann Labb{\'e}$^{4}$\namesep
 Gu Wang$^{5}$\\
 Eric Brachmann$^{6}$\namesep
 Bertram Drost$^{7}$\namesep
 Carsten Rother$^{8}$\namesep
 Ji{\v{r}}{\'i}~Matas$^{9}$\vspace{0.7em} \\
 {\normalsize
     {$^{1}$German Aerospace Center}\namesep
     {$^{2}$TU Munich}\namesep
     {$^{3}$Reality Labs at Meta}\namesep
     {$^{4}$INRIA Paris} \namesep
     {$^{5}$Tsinghua University}
 }\\
 {\normalsize
     {$^{6}$Niantic}\namesep
     {$^{7}$MVTec}\namesep
     {$^{8}$Heidelberg University}\namesep
     {$^{9}$Czech Technical University in Prague}
 }
 \vspace{0.7ex}
}

\maketitle

\begin{abstract}
We present the evaluation methodology, datasets and results of the BOP Challenge 2022, the fourth in a series of public competitions organized with the goal to capture the status quo in the field of 6D object pose estimation from an RGB/RGB-D image.
In 2022, we witnessed another significant improvement in the pose estimation accuracy -- the state of the art, which was 56.9~AR$_C$ in 2019 (Vidal et al.) and 69.8~AR$_C$ in 2020 (CosyPose), moved to new heights of 83.7~AR$_C$ (GDRNPP). Out of 49 pose estimation methods evaluated since 2019, the top 18 are from 2022.
Methods based on point pair features, which were introduced in 2010 and achieved competitive results even in 2020, are now clearly outperformed by deep learning methods.
The synthetic-to-real domain gap was again significantly reduced, with 82.7 AR$_C$ achieved by GDRNPP trained only on synthetic images from BlenderProc. The fastest variant of GDRNPP reached 80.5 AR$_C$ with an average time per image of 0.23s.
Since most of the recent methods for 6D object pose estimation begin by detecting/segmenting objects, we also started evaluating 2D object detection and segmentation performance based on the COCO metrics.
Compared to the Mask R-CNN results from CosyPose in 2020, detection improved from 60.3 to 77.3~AP$_C$ and segmentation from 40.5 to 58.7~AP$_C$.
The online evaluation system stays open and is available at:
\texttt{\href{http://bop.felk.cvut.cz/}{bop.felk.cvut.cz}}.
\end{abstract}

\begin{figure*}[t!]
\begin{center}
    \begingroup
    \setlength{\tabcolsep}{0.0pt}
    \renewcommand{\arraystretch}{1.0}
    \begin{tabular}{ c c c c }
        \includegraphics[width=0.244\linewidth]{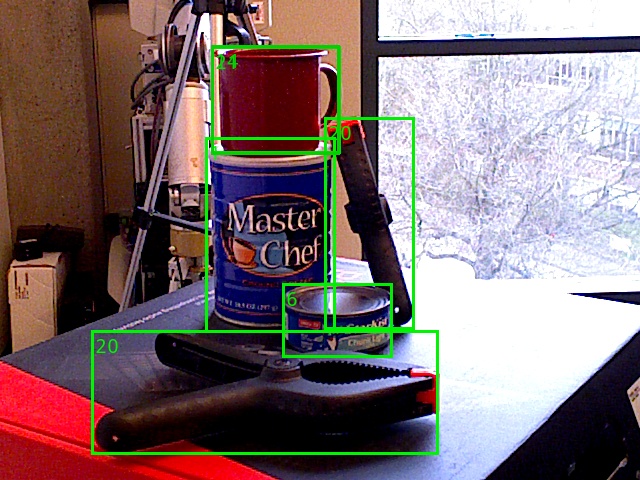} \hspace{0.25ex} &
        \includegraphics[width=0.244\linewidth]{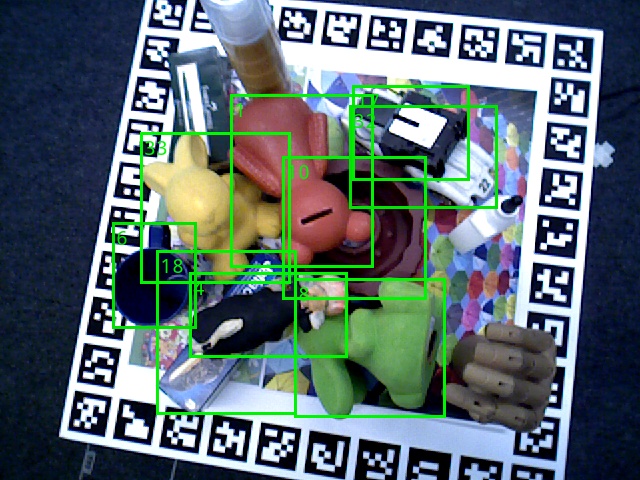} \hspace{0.25ex} &
        \includegraphics[width=0.244\linewidth]{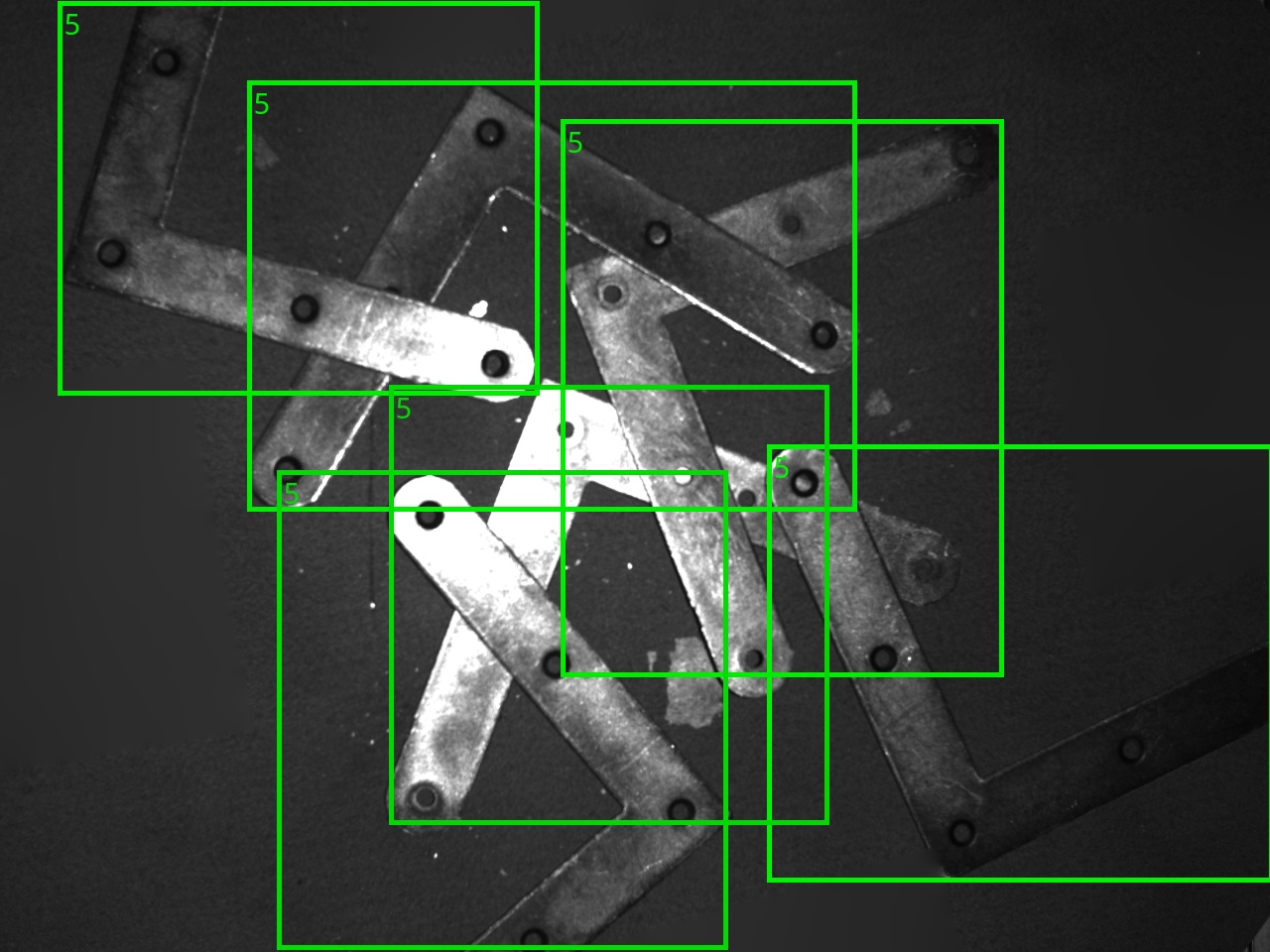} \hspace{0.25ex} &
        \includegraphics[width=0.244\linewidth]{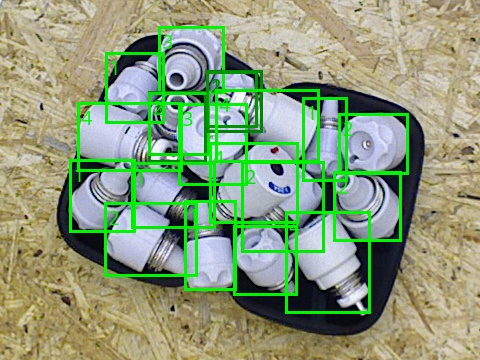} \\

        \includegraphics[width=0.244\linewidth]{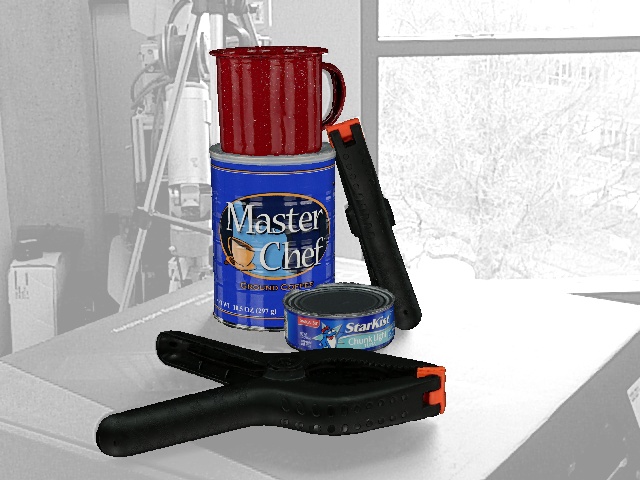} \hspace{0.25ex} &
        \includegraphics[width=0.244\linewidth]{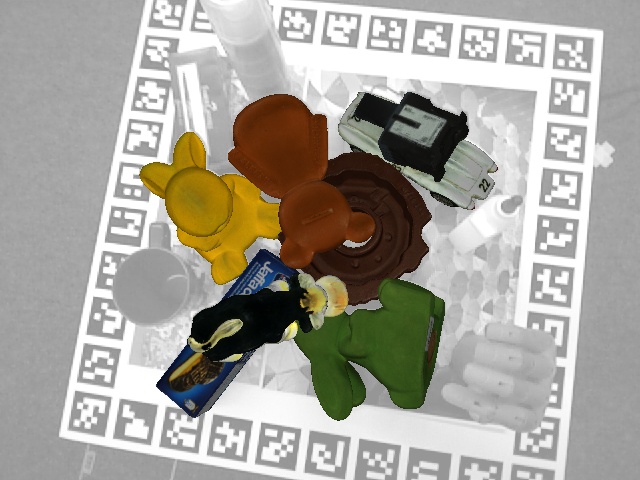} \hspace{0.25ex} &
        \includegraphics[width=0.244\linewidth]{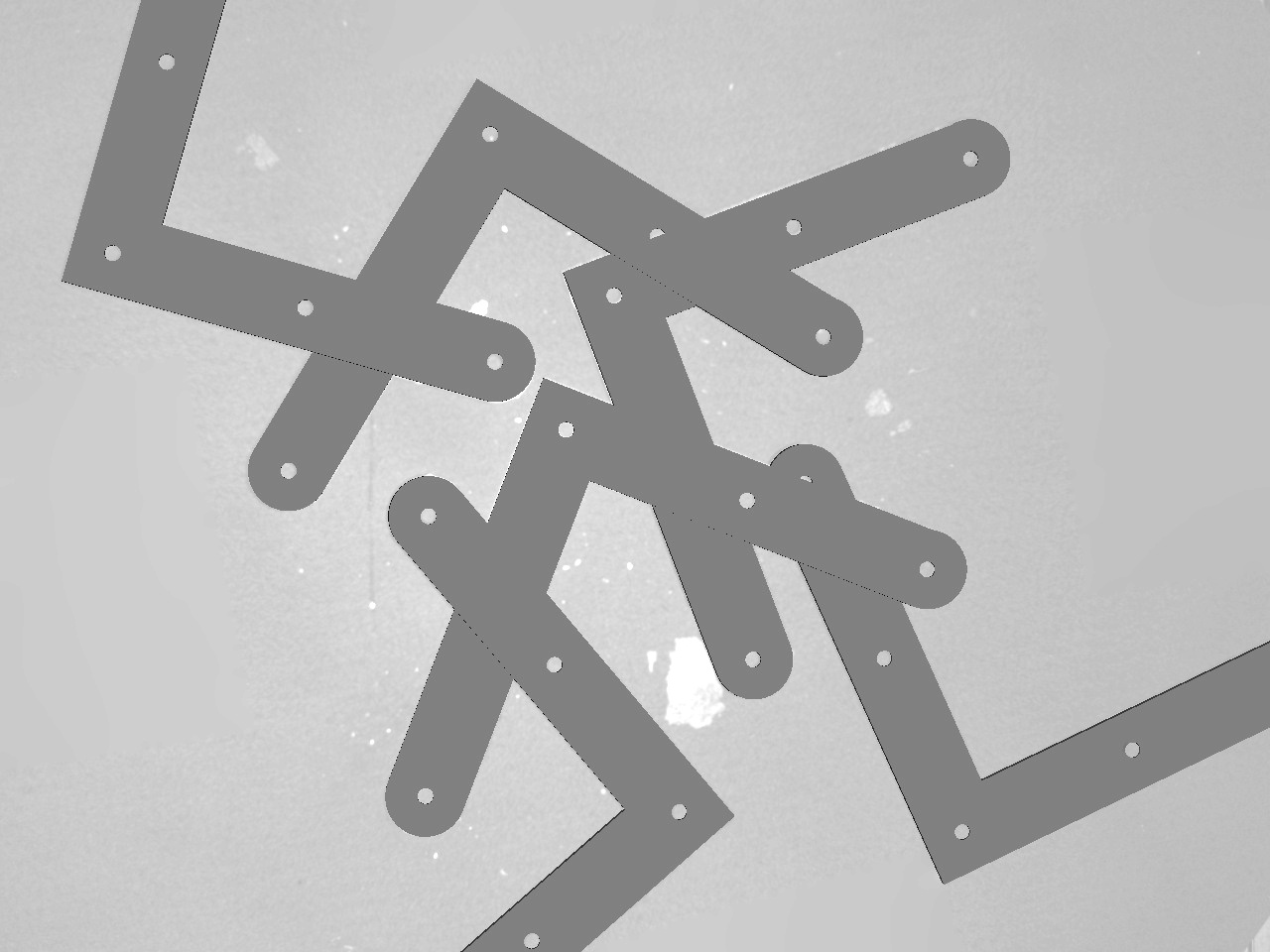} \hspace{0.25ex} &
        \includegraphics[width=0.244\linewidth]{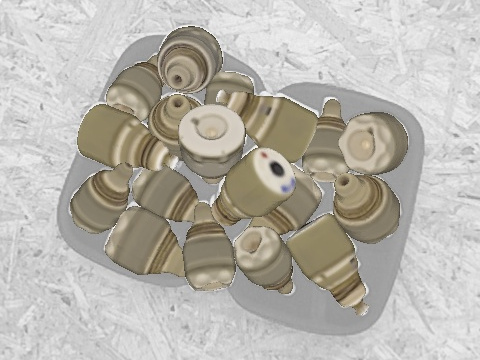}
    \end{tabular}
    \endgroup
    \caption{
    \textbf{2D object detection followed by 6D pose estimation from the detected regions} is a strategy used by the majority of recent 6D object pose estimation methods.
    This figure shows detections (top) and 3D object models rendered in estimated poses (bottom) produced by the 2022 top-performing method, 
    GDRNPP~\cite{Wang_2021_GDRN,liu2022gdrnpp_bop}, on challenging images from
    YCB-V~\cite{xiang2017posecnn}, HB~\cite{kaskman2019homebreweddb}, ITODD~\cite{drost2017introducing}, and T-LESS~\cite{hodan2017tless}.
    }
    \label{fig:cover_results}
\end{center}
\end{figure*}

\section{Introduction}

Estimating the 6D pose, \ie, the 3D translation and 3D rotation, of specific rigid objects from a single image is an important task for application fields such as robotic manipulation, augmented reality, or autonomous driving.
The BOP Challenge 2022 is the fourth in a series of public challenges that are part of the BOP\footnote{BOP stands for Benchmark for 6D Object Pose Estimation~\cite{hodan2018bop}.} project aiming to continuously report the state of the art in 6D object pose estimation.
The first challenge was organized in 2017~\cite{hodan2017sixd} and the results were published in~\cite{hodan2018bop}.
Results of the second challenge from 2019~\cite{hodan2019bop}, the third from 2020 \cite{hodan2020bop}, and the fourth from 2022 are included and discussed in this paper.

Participants of the 2022 challenge were competing on three tasks: 6D object localization, 2D object detection, and 2D object segmentation.
The 6D object localization task has the same evaluation methodology and leaderboard since 2019, while the latter two tasks were introduced in 2022.

In the 6D object localization task, methods report their predictions on the basis of two sources of information. Firstly, at training time, a method is given 3D object models and training images showing the objects in known 6D poses. Secondly, at test time, the method is provided with a test image and a list of object instances visible in the image, and the goal is to estimate 6D poses of the listed instances. The images consist of RGB-D (aligned color and depth) channels and intrinsic camera parameters are known.

The 2D object detection and segmentation tasks were introduced to address the design of the majority of recent object pose estimation methods, which start by detecting/segmenting objects and then estimate their poses from the predicted image regions. Evaluating the detection/segmentation and pose estimation stages separately enables a better understanding of advances in the two stages.
To create an opportunity for detector-agnostic comparison of pose estimation methods and to allow participants to focus only on the pose estimation stage, we also provided default detections and segmentations from Mask R-CNN~\cite{he2017mask} trained for CosyPose~\cite{labbe2020cosypose}, the winning method in 2020.

The challenge primarily focuses on the practical scenario where no real images are available at training time, only the 3D object models and images synthesized using the models. While capturing real images of objects under various conditions and annotating the images with 6D object poses requires a significant human effort~\cite{hodan2017tless}, the 3D models are either available before the physical objects, which is often the case for manufactured objects, or can be reconstructed at an admissible cost.
Approaches for reconstructing 3D models of opaque, matte and moderately specular objects are established~\cite{newcombe2011kinectfusion,reizenstein2021common} and promising approaches for transparent and highly specular objects are emerging~\cite{wu2018full,Munkberg_2022_CVPR,hasselgren2022shape,verbin2022ref}.

In the 2019 challenge, methods using the depth image channel were mostly based on point pair features (PPF's)~\cite{drost2010model} and clearly outperformed methods relying only on the RGB channels, all of which were based on deep neural networks (DNN's). DNN-based methods need large amounts of annotated training images, which had been typically obtained by OpenGL rendering of the 3D object models on random backgrounds~\cite{kehl2017ssd,hinterstoisser2017pre}. However, as suggested in~\cite{hodan2019photorealistic}, the evident domain gap between these ``render\;\&\;paste'' training images and real test images
limits the potential of the DNN-based methods. To reduce the gap between the synthetic and real domains and thus to bring fresh air to the DNN world, we joined the development of BlenderProc\footnote{\href{https://github.com/DLR-RM/BlenderProc/blob/main/README_BlenderProc4BOP.md}{\texttt{github.com/DLR-RM/BlenderProc}}}~\cite{denninger2019blenderproc,denninger2020blenderproc}, an open-source, physically-based renderer (PBR).
For the 2020 challenge, we then provided participants with 350K
PBR training images (see~\cite{hodan2020bop} for examples), which helped the DNN-based methods to achieve noticeably higher accuracy
and to finally catch up with the PPF-based methods.

In the 2022 challenge, DNN-based methods for 6D object localization clearly outperformed PPF-based methods in both accuracy and speed, with the performance gains coming mostly from advances in network architectures and training schemes. The largest improvements were achieved on challenging industry-relevant datasets ITODD~\cite{drost2017introducing} and T-LESS~\cite{hodan2017tless}, and on the HB dataset~\cite{kaskman2019homebreweddb} which includes diverse objects captured under various levels of occlusion. Remarkably, RGB methods from 2022 surpassed RGB-D methods from 2020, the performance gap between methods trained only on PBR images and methods trained also on real images noticeably shrinked, and some methods started training on the depth image channel in addition to the RGB channels. On the new 2D object detection and segmentation tasks, large gains were achieved \wrt a baseline from 2020.

Sec.~\ref{sec:methodology} of this paper defines the evaluation methodology, Sec.~\ref{sec:datasets} introduces datasets,
Sec.~\ref{sec:evaluation} describes the experimental setup and analyzes the results, Sec.~\ref{sec:awards} presents the awards of the BOP Challenge 2022, and Sec.~\ref{sec:conclusion} concludes the paper.

\section{Evaluation Methodology} \label{sec:methodology}

Methods are evaluated on the task of 6D object localization, as in 2019 and 2020~\cite{hodan2020bop}, and additionally on the tasks of 2D object detection and 2D object segmentation. The tasks are defined below together with accuracy scores that are used to compare methods. Participants could submit their results to any of the three tasks. Note that although all BOP datasets currently include RGB-D images (Sec.~\ref{sec:datasets}), a method may have used any of the image channels.

\subsection{2D Object Detection and Segmentation Tasks}
\label{sec:2d_tasks}

\noindent\textbf{Training input:}
At training time, a detection/segmentation method is provided a set of training images showing objects annotated with ground-truth 2D bounding boxes (for the detection task) and binary masks (for the segmentation task). The boxes are \emph{amodal} (covering the whole object silhouette, including the occluded parts) while the masks are \emph{modal} (covering only the visible object part). The method can also use 3D mesh models that are available for the objects (\eg, to synthesize extra training images).

\customparagraph{Test input:}
At test time, the method is given an image showing an arbitrary number of instances of an arbitrary number of objects from a considered dataset. No prior information about the visible object instances is provided.

\customparagraph{Test output:} The method produces a list of
amodal 2D bounding boxes (for detection) and modal binary masks (for segmentation) with confidences.

\customparagraph{Metrics:} Following the the evaluation methodology from the COCO 2020 Object Detection Challenge~\cite{lin2014microsoft}, the detection/segmentation accuracy is measured by the Average Precision (AP). Specifically, a per-object $\text{AP}_O$ score is calculated by averaging the precision at multiple Intersection over Union (IoU) thresholds: $[0.5, 0.55, \dots , 0.95]$. The accuracy of a method on a dataset $D$ is measured by $\text{AP}_D$ calculated by averaging per-object $\text{AP}_O$ scores, and the overall accuracy on the core datasets (Sec.~\ref{sec:datasets}) is measured by $\text{AP}_C$ defined as the average of the per-dataset $\text{AP}_D$ scores.

Analagous to the 6D localization task, only object instances for which at least $10\%$ of the projected surface area is visible need to be detected/segmented. Correct predictions for objects that are visible from less than $10\%$ are filtered out and not counted as false positives. Up to $100$ predictions with the highest scores per image are considered.

\subsection{6D Object Localization Task}
\label{sec:6d_localization_task}

As in the 2019 and 2020 editions of the challenge, methods are evaluated on the task of 6D localization of a \textbf{v}arying number of \textbf{i}nstances of a \textbf{v}arying number of \textbf{o}bjects from a single image. This variant of the 6D object localization task is referred to as ViVo and defined as follows.\footnote{See Sec.~A.1 in~\cite{hodan2020bop} for a discussion on why the methods are evaluated on 6D object localization instead of 6D object detection, where no prior information about the visible object instances is provided~\cite{hodan2016evaluation}.}

\customparagraph{Training input:}
A method is provided a set of training images showing objects annotated with 6D poses, and 3D mesh models of the objects (typically with a color texture).
A 6D pose is defined by a matrix $\textbf{P} = [\mathbf{R} \, | \, \mathbf{t}]$, where $\mathbf{R}$ is a 3D rotation matrix, and $\mathbf{t}$ is a 3D translation vector. The matrix $\textbf{P}$ defines a rigid transformation from the 3D space of the object model to the 3D space of the camera.

\customparagraph{Test input:}
The method is given an image unseen during training and a list $L = [(o_1, n_1),$ $\dots,$ $(o_m, n_m)]$, where $n_i$ is the number of instances of object $o_i$ visible in the image.

\customparagraph{Test output:} The method outputs a list $E=[E_1,$$\dots,$$E_m]$, where $E_i$ is a list of $n_i$ pose estimates with confidences for instances of object $o_i$.

\customparagraph{Metrics:}
The 6D object localization task is evaluated as in the 2020 challenge~\cite{hodan2020bop}.
In short, the error of an estimated pose \wrt the ground-truth pose is calculated by three pose-error functions: Visible Surface Discrepancy (VSD) which treats indistinguishable poses as equivalent by considering only the visible object part, Maximum Symmetry-Aware Surface Distance (MSSD) which considers a set of pre-identified global object symmetries and measures the surface deviation in 3D, and Maximum Symmetry-Aware Projection Distance (MSPD) which considers the object symmetries and measures the perceivable deviation.
An estimated pose is considered correct \wrt a pose-error function~$e$, if $e < \theta_e$, where $e \in \{\text{VSD}, \text{MSSD}, \text{MSPD}\}$ and $\theta_e$ is the threshold of correctness. The fraction of annotated object instances for which a correct pose is estimated is referred to as Recall. The Average Recall \wrt a function~$e$, denoted as $\text{AR}_e$, is defined as the average of the Recall rates calculated for multiple settings of the threshold $\theta_e$ and also for multiple settings of a misalignment tolerance $\tau$ in the case of $\text{VSD}$. The accuracy of a method on a dataset $D$ is measured by: $\text{AR}_D = (\text{AR}_\text{VSD} + \text{AR}_{\text{MSSD}} + \text{AR}_{\text{MSPD}}) \, / \, 3$, which is calculated over estimated poses of all objects from $D$. The overall accuracy on the core datasets is measured by $\text{AR}_C$ defined as the average of the per-dataset $\text{AR}_D$ scores.\footnote{When calculating AR$_C$, scores are not averaged over objects before averaging over datasets, which is done when calculating $\text{AP}_C$ (Sec.~\ref{sec:2d_tasks}) to comply with the original COCO evaluation methodology~\cite{lin2014microsoft}.}

\begin{figure}[h!]
\begin{center}

\begingroup
\footnotesize

\renewcommand{\arraystretch}{0.9}

\begin{tabular}{ @{}c@{ } @{}c@{ } @{}c@{ } @{}c@{ } }
LM~\cite{hinterstoisser2012accv} & LM-O*~\cite{brachmann2014learning} & T-LESS*~\cite{hodan2017tless} & ITODD*~\cite{drost2017introducing} \vspace{0.5ex} \\
\includegraphics[width=0.243\columnwidth]{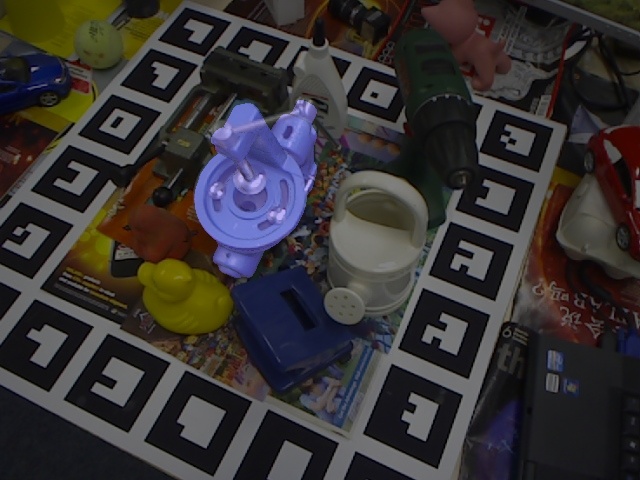} &
\includegraphics[width=0.243\columnwidth]{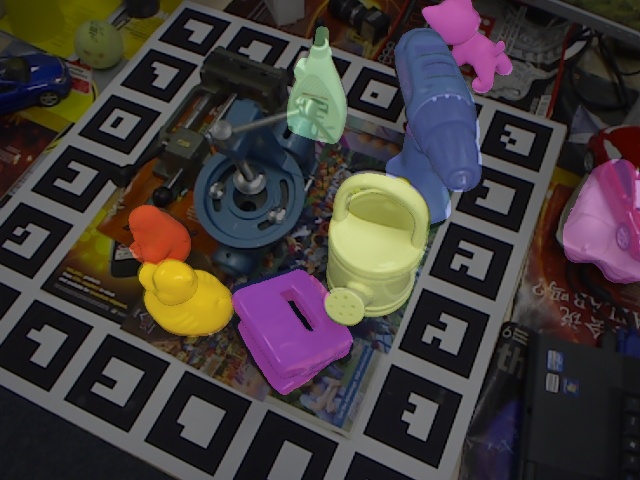} &
\includegraphics[width=0.243\columnwidth]{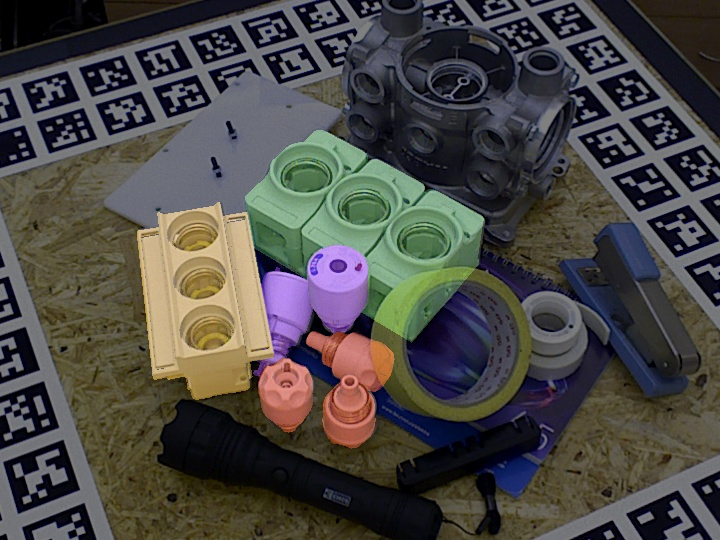} &
\includegraphics[width=0.243\columnwidth]{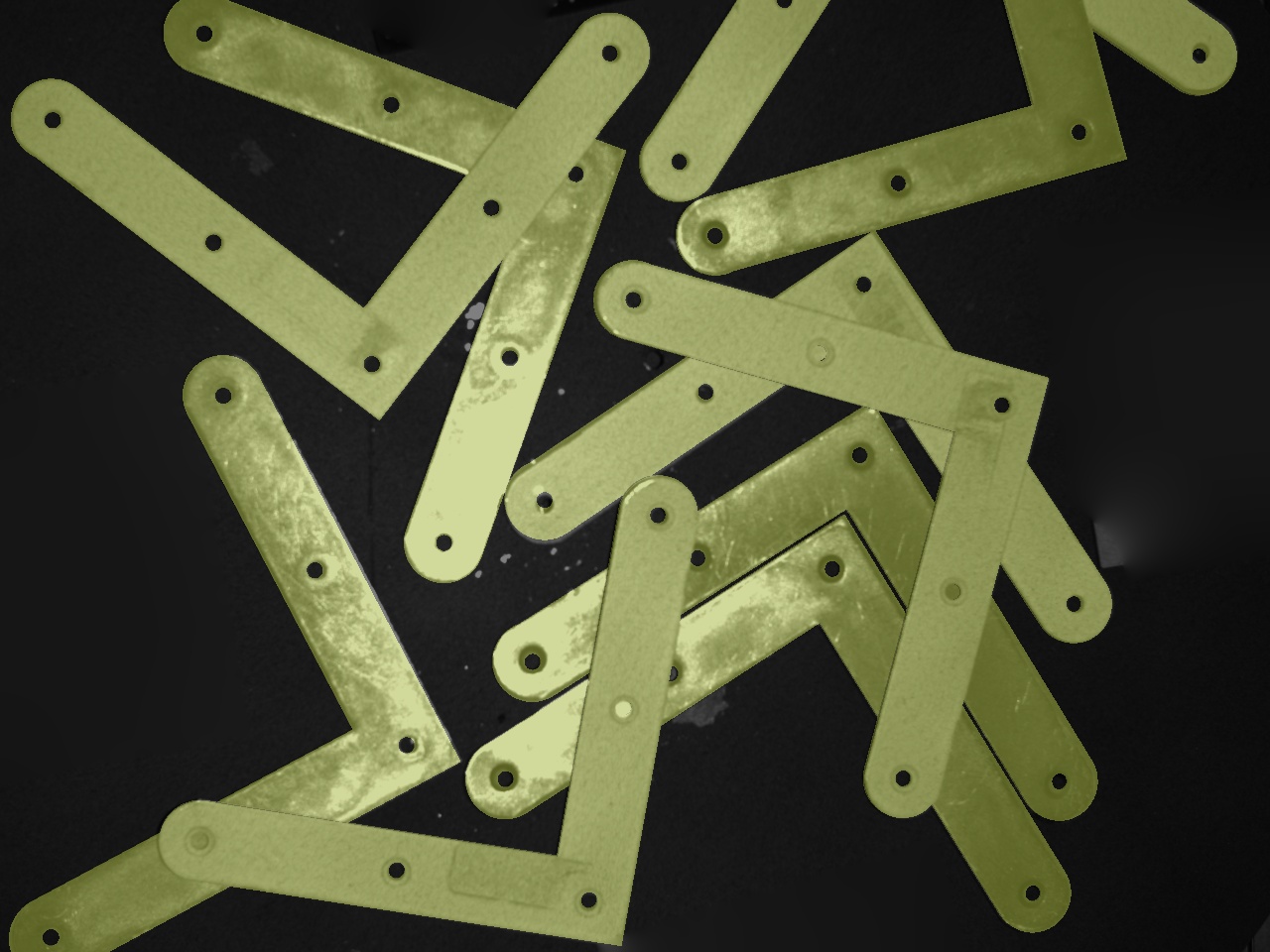} \vspace{0.1ex} \\
\end{tabular}

\begin{tabular}{ @{}c@{ } @{}c@{ } @{}c@{ } @{}c@{ } }
HB*~\cite{kaskman2019homebreweddb} & YCB-V*~\cite{xiang2017posecnn} & RU-APC~\cite{rennie2016dataset} & IC-BIN*~\cite{doumanoglou2016recovering} \vspace{0.5ex} \\
\includegraphics[width=0.243\columnwidth]{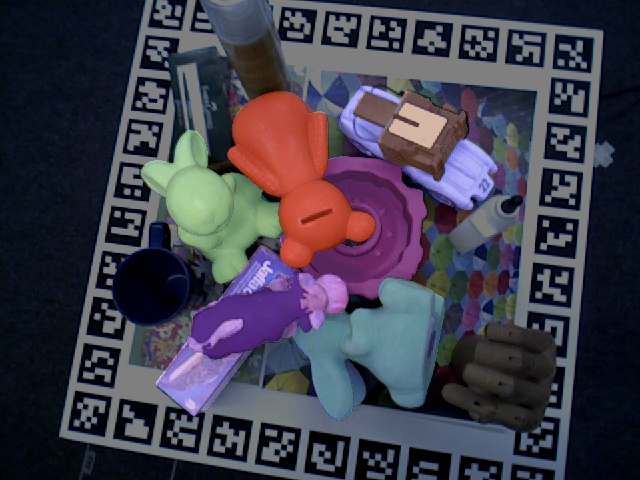} &
\includegraphics[width=0.243\columnwidth]{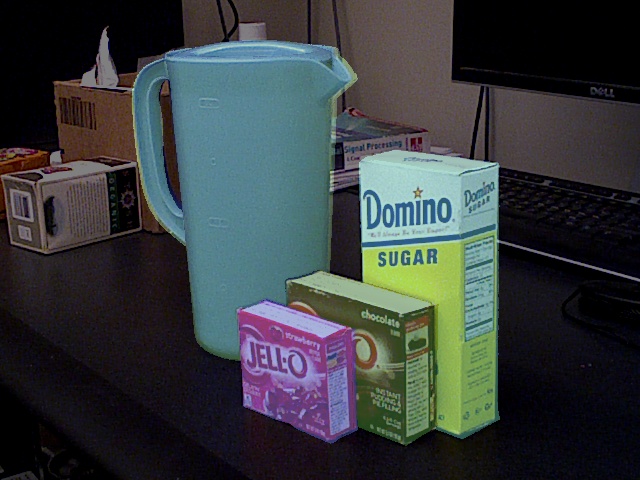} &
\includegraphics[width=0.243\columnwidth]{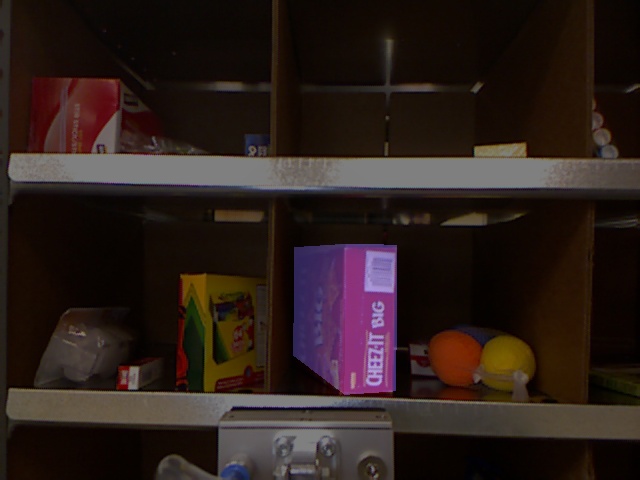} &
\includegraphics[width=0.243\columnwidth]{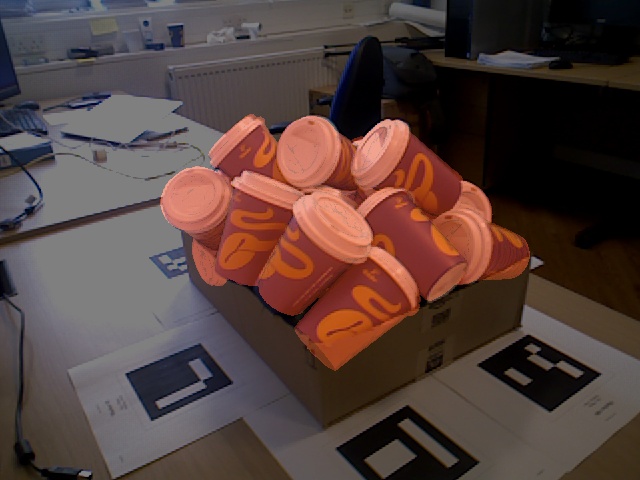} \vspace{0.1ex} \\
\end{tabular}

\begin{tabular}{ @{}c@{ } @{}c@{ } @{}c@{ } @{}c@{ }}
IC-MI~\cite{tejani2014latent} & TUD-L*~\cite{hodan2018bop} & TYO-L~\cite{hodan2018bop} & HOPE~\cite{tyree2022hope}\vspace{0.5ex} \\
\includegraphics[width=0.243\columnwidth]{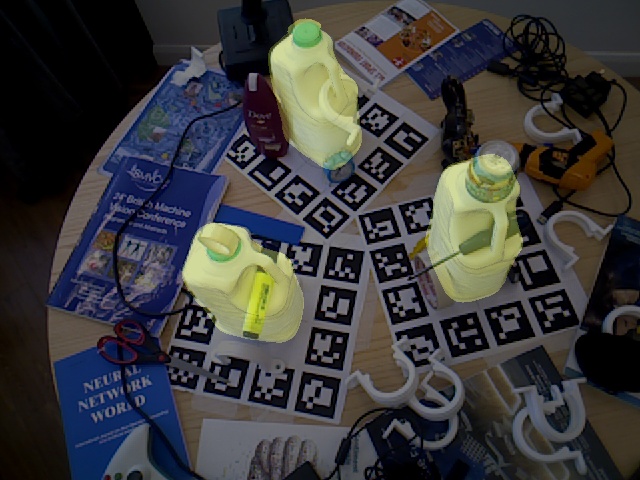} &
\includegraphics[width=0.243\columnwidth]{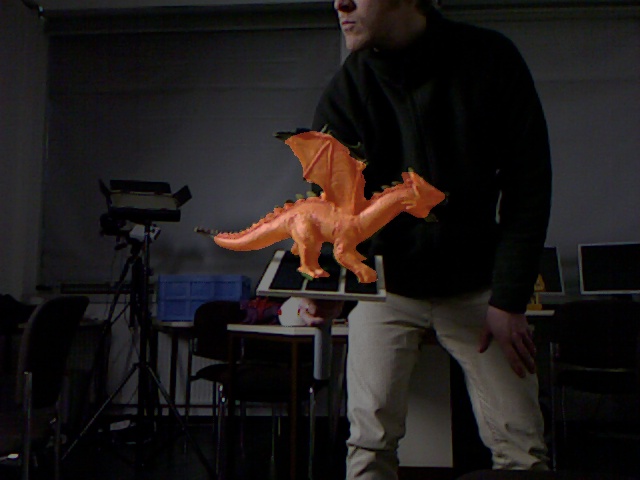} &
\includegraphics[width=0.243\columnwidth]{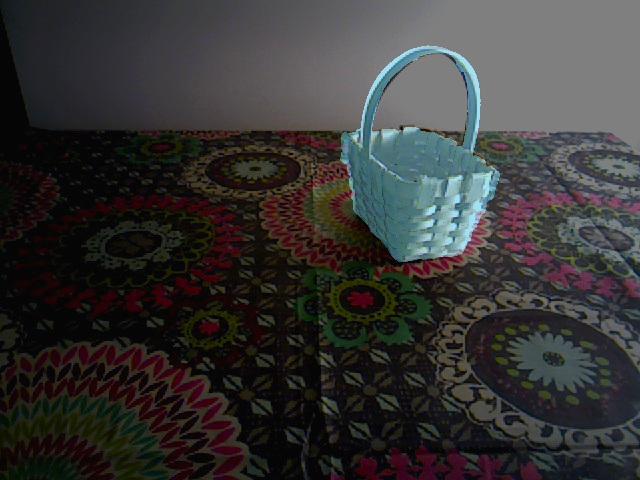} &
\includegraphics[width=0.243\columnwidth]{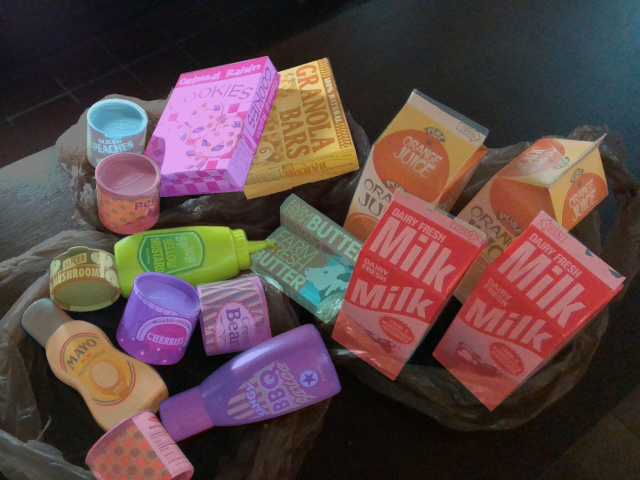} \\
\end{tabular}

\endgroup

\caption{\label{fig:dataset_overview}
\textbf{An overview of the BOP datasets.} The seven core datasets are marked with a star. Shown are RGB channels of sample test images which were darkened and overlaid with colored 3D object models in the ground-truth 6D poses.
\vspace{4ex}
}

\begingroup
\setlength\tabcolsep{1.0pt}
\footnotesize
\begin{tabularx}{\columnwidth}{ l r R R R R R R R }
	\toprule
	&
	&
	\multicolumn{2}{c}{Train. im.} &
	\multicolumn{1}{c}{Val im.} &
	\multicolumn{2}{c}{Test im.} &
	\multicolumn{2}{c}{Test inst.} \\
	\cmidrule(l{2pt}r{2pt}){3-4} \cmidrule(l{2pt}r{2pt}){5-5} \cmidrule(l{2pt}r{2pt}){6-7} \cmidrule(l{2pt}r{2pt}){8-9}
	\multicolumn{1}{l}{Dataset} &
	\multicolumn{1}{c}{Obj.} &
	\multicolumn{1}{c}{Real} &
	\multicolumn{1}{c}{PBR} &
	\multicolumn{1}{c}{Real} &
	\multicolumn{1}{c}{All} &
	\multicolumn{1}{c}{Used} &
	\multicolumn{1}{c}{All} &
	\multicolumn{1}{c}{Used} \\
	\midrule

	LM-O \cite{brachmann2014learning} & 8 & -- & 50K & -- & 1214 & 200 & 9038 & 1445 \\
	T-LESS \cite{hodan2017tless} &  30 & 37584 & 50K & -- & 10080 & 1000 & 67308 & 6423 \\
	ITODD \cite{drost2017introducing} &  28 & -- & 50K & 54 & 721 & 721 & 3041 & 3041 \\
	HB \cite{kaskman2019homebreweddb} &  33 & -- & 50K & 4420 & 13000 & 300 & 67542 & 1630 \\
	YCB-V \cite{xiang2017posecnn} &  21 & 113198 & 50K & -- & 20738 & 900 & 98547 & 4123 \\
	TUD-L \cite{hodan2018bop} &  3 & 38288 & 50K & -- & 23914 & 600 & 23914 & 600 \\
	IC-BIN \cite{doumanoglou2016recovering} & 2 & -- & 50K & -- & 177 & 150 & 2176 & 1786 \\
	\midrule
	LM \cite{hinterstoisser2012accv} & 15 & -- & 50K & -- & 18273 & 3000 & 18273 & 3000 \\
	RU-APC \cite{rennie2016dataset} & 14 & -- & -- & -- & 5964 & 1380 & 5964 & 1380\\
	IC-MI \cite{tejani2014latent} & 6 & -- & -- & -- & 2067 & 300 & 5318 & 800 \\
	TYO-L \cite{hodan2018bop} & 21 & -- & -- & -- & 1670 & 1670 & 1670 & 1670 \\
        HOPE \cite{tyree2022hope} & 28 & -- & -- & 50 & 188 & 188 & 3472 & 2898 \\
	\bottomrule
\end{tabularx}

\endgroup

\captionof{table}{\label{tab:dataset_params} \textbf{Parameters of the BOP datasets.} The core datasets are listed in the upper part.
PBR training images rendered by BlenderProc~\cite{denninger2019blenderproc,denninger2020blenderproc} are provided for all core datasets.
Most datasets include also OpenGL-rendered training images of 3D object models on a black background (not shown in the table).
If a dataset includes both validation and test images, ground-truth annotations are public only for the validation images. All test images are real.
Column ``Test inst./All'' shows the number of annotated object instances for which at least $10\%$ of the projected surface area is visible in the test image. Columns ``Used'' show the number of test images and object instances used in the BOP Challenge 2019, 2020, and 2022.
}
\vspace{-1.7ex}
\end{center}
\end{figure}

\section{Datasets} \label{sec:datasets}

BOP currently includes twelve datasets in a unified format -- sample test images are in Fig.~\ref{fig:dataset_overview} and dataset parameters in Tab.~\ref{tab:dataset_params}. Seven from the twelve were selected as core datasets:
LM-O, T-LESS, ITODD, HB, YCB-V, TUD-L, IC-BIN.
A method had to be evaluated on all core datasets to be considered for the main challenge awards (Sec.~\ref{sec:awards}).

Each dataset includes 3D object models and training and test RGB-D images annotated with ground-truth 6D object poses. The object models are provided in the form of 3D meshes (in most cases with a color texture) which were created manually or using KinectFusion-like systems for 3D reconstruction~\cite{newcombe2011kinectfusion}. While all test images are real, training images may be real and/or synthetic.
The seven core datasets include a total of 350K photorealistic PBR (physically-based rendered) training images generated and automatically annotated using BlenderProc~\cite{denninger2019blenderproc,denninger2020blenderproc}.
Example images are shown in \cite{hodan2020bop} and a detailed description of the generation process and an analysis of the importance of PBR training images is provided in Sec. 3.2 and 4.3 of the 2020 challenge paper~\cite{hodan2020bop}. Datasets T-LESS, TUD-L and YCB-V include also real training images, and most datasets additionally include training images obtained by OpenGL rendering of the 3D object models on a black background.
Test images were captured in scenes with graded complexity, often with clutter and occlusion. The HB and ITODD datasets include also real validation images -- in this case, the ground-truth poses are publicly available only for the validation and not for the test images.
The datasets can be downloaded from the BOP website\footnote{\texttt{\href{http://bop.felk.cvut.cz/datasets}{bop.felk.cvut.cz/datasets}}} and more details about the datasets can be found in Chapter 7 of~\cite{hodan2021phd}.

\section{Results and Discussion} \label{sec:evaluation}

This section presents results of the BOP Challenge 2022, compares them with results from 2019 and 2020 challenge editions, and summarizes the main messages for our field.

In total, 49 methods were evaluated on the ViVo variant of the 6D object localization task on all seven core datasets -- 11 methods in 2019, 15 in 2020, and 23 in 2022. Additionally, 8 methods were evaluated on the new detection task and 8 methods on the new segmentation task.

\subsection{Experimental Setup}

Participants of the BOP Challenge 2022 were submitting results of their methods to the online evaluation system at \texttt{\href{http://bop.felk.cvut.cz/}{bop.felk.cvut.cz}} from May 1, 2022 until the deadline on October 16, 2022. The methods were evaluated on the ViVo variant of the 6D object localization task as described in Sec.~\ref{sec:6d_localization_task} and on the 2D object detection and segmentation tasks as described in Sec.~\ref{sec:2d_tasks}. The evaluation scripts are publicly available in the BOP toolkit.\footnote{\texttt{\href{https://github.com/thodan/bop_toolkit}{github.com/thodan/bop\_toolkit}}}

A method had to use a fixed set of hyper-parameters across all objects and datasets. For training, a method may have used the provided object models and training images, and rendered extra training images using the object models. However, not a single pixel of test images may have been used for training, nor the individual ground-truth poses or object masks provided for the test images.
Ranges of the azimuth and elevation camera angles, and a range of the camera-object distances determined by the ground-truth poses from test images is the only information about the test set that may have been used during training.

Only subsets of test images were used to remove redundancies and speed up the evaluation, and only object instances for which at least $10\%$ of the projected surface area is visible were considered in the evaluation.

\setlength{\tabcolsep}{2pt}
\begin{table*}[t]
    \footnotesize
    \centering
    \begin{tabularx}{\linewidth}{rlYYYYYYYYY}
        \toprule
        \# & Method & LM-O & T-LESS & TUD-L & IC-BIN & ITODD & HB & YCB-V & \mbox{$\text{AR}_C$} & Time \\ 
        \midrule
        1 & GDRNPP-PBRReal-RGBD-MModel \cite{Wang_2021_GDRN,liu2022gdrnpp_bop}& \cellcolor{arcol!77.5}77.5	&\cellcolor{arcol!87.4}87.4	&\cellcolor{arcol!96.6}96.6	&\cellcolor{arcol!72.2}72.2&	\cellcolor{arcol!67.9}67.9&	\cellcolor{arcol!92.6}92.6&	\cellcolor{arcol!92.1}92.1& \cellcolor{avgcol!83.7}83.7 & \cellcolor{timecol!62.63}$\phantom{0}$6.26 \\ 
        2 & GDRNPP-PBR-RGBD-MModel \cite{Wang_2021_GDRN,liu2022gdrnpp_bop}& \cellcolor{arcol!77.5}77.5	&\cellcolor{arcol!85.2}85.2	&\cellcolor{arcol!92.9}92.9	&\cellcolor{arcol!72.2}72.2&	\cellcolor{arcol!67.9}67.9&	\cellcolor{arcol!92.6}92.6&	\cellcolor{arcol!90.6}90.6& \cellcolor{avgcol!82.7}82.7 & \cellcolor{timecol!62.64}$\phantom{0}$6.26 \\ 
        3 & GDRNPP-PBRReal-RGBD-MModel-Fast \cite{Wang_2021_GDRN,liu2022gdrnpp_bop}& \cellcolor{arcol!79.2}79.2	&\cellcolor{arcol!87.2}87.2	&\cellcolor{arcol!93.6}93.6	&\cellcolor{arcol!70.2}70.2&	\cellcolor{arcol!58.8}58.8&	\cellcolor{arcol!90.9}90.9&	\cellcolor{arcol!83.4}83.4& \cellcolor{avgcol!80.5}80.5 & \cellcolor{timecol!02.28}$\phantom{0}$0.23 \\ 
        4 & GDRNPP-PBRReal-RGBD-MModel-Offi. \cite{Wang_2021_GDRN,liu2022gdrnpp_bop} & \cellcolor{arcol!75.8}75.8	&\cellcolor{arcol!82.4}82.4	&\cellcolor{arcol!96.6}96.6	&\cellcolor{arcol!70.8}70.8&	\cellcolor{arcol!54.3}54.3&	\cellcolor{arcol!89.0}89.0&	\cellcolor{arcol!89.6}89.6& \cellcolor{avgcol!79.8}79.8 & \cellcolor{timecol!64.06}$\phantom{0}$6.41 \\
        5 & Extended\_FCOS+PFA-MixPBR-RGBD~\cite{hu2022perspective} & \cellcolor{arcol!79.7}79.7	&\cellcolor{arcol!85.0}85.0	&\cellcolor{arcol!96.0}96.0	&\cellcolor{arcol!67.6}67.6&	\cellcolor{arcol!46.9}46.9&	\cellcolor{arcol!86.9}86.9&	\cellcolor{arcol!88.8}88.8& \cellcolor{avgcol!78.7}78.7 & \cellcolor{timecol!23.17}$\phantom{0}$2.32 \\ 
        6 & Extended\_FCOS+PFA-MixPBR-RGBD-Fast~\cite{hu2022perspective} & \cellcolor{arcol!79.2}79.2	&\cellcolor{arcol!77.9}77.9	&\cellcolor{arcol!95.8}95.8	&\cellcolor{arcol!67.1}67.1&	\cellcolor{arcol!46.0}46.0&	\cellcolor{arcol!86.0}86.0&	\cellcolor{arcol!88.0}88.0& \cellcolor{avgcol!77.1}77.1 & \cellcolor{timecol!06.39}$\phantom{0}$0.64 \\ 
        7 & RCVPose3D-SingleModel-VIVO-PBR~\cite{wu2022keypoint} & \cellcolor{arcol!72.9}72.9	&\cellcolor{arcol!70.8}70.8	&\cellcolor{arcol!96.6}96.6	&\cellcolor{arcol!73.3}73.3&	\cellcolor{arcol!53.6}53.6&	\cellcolor{arcol!86.3}86.3&	\cellcolor{arcol!84.3}84.3& \cellcolor{avgcol!76.8}76.8 & \cellcolor{timecol!13.36}$\phantom{0}$1.34 \\ 
        8 & ZebraPoseSAT-EffnetB4+ICP(DefaultDet)~\cite{su2022zebrapose} & \cellcolor{arcol!75.2}75.2	&\cellcolor{arcol!72.7}72.7	&\cellcolor{arcol!94.8}94.8	&\cellcolor{arcol!65.2}65.2&	\cellcolor{arcol!52.7}52.7&	\cellcolor{arcol!88.3}88.3&	\cellcolor{arcol!86.6}86.6& \cellcolor{avgcol!76.5}76.5 & \cellcolor{timecol!05.00}$\phantom{0}$0.50 \\ 
        9 & Extended\_FCOS+PFA-PBR-RGBD~\cite{hu2022perspective} & \cellcolor{arcol!79.7}79.7	&\cellcolor{arcol!80.2}80.2	&\cellcolor{arcol!89.3}89.3	&\cellcolor{arcol!67.6}67.6&	\cellcolor{arcol!46.9}46.9&	\cellcolor{arcol!86.9}86.9&	\cellcolor{arcol!82.6}82.6& \cellcolor{avgcol!76.2}76.2 & \cellcolor{timecol!26.31}$\phantom{0}$2.63 \\ 
        10 & SurfEmb-PBR-RGBD~\cite{haugaard2022surfemb} & \cellcolor{arcol!76.0}76.0	&\cellcolor{arcol!82.8}82.8	&\cellcolor{arcol!85.4}85.4	&\cellcolor{arcol!65.9}65.9&	\cellcolor{arcol!53.8}53.8&	\cellcolor{arcol!86.6}86.6&	\cellcolor{arcol!79.9}79.9& \cellcolor{avgcol!75.8}75.8 & \cellcolor{timecol!90.48}$\phantom{0}$9.05 \\ 
        11 & GDRNPP-PBRReal-RGBD-SModel \cite{Wang_2021_GDRN,liu2022gdrnpp_bop} & \cellcolor{arcol!75.7}75.7	&\cellcolor{arcol!85.6}85.6	&\cellcolor{arcol!90.6}90.6	&\cellcolor{arcol!68.0}68.0&	\cellcolor{arcol!35.6}35.6&	\cellcolor{arcol!86.4}86.4&	\cellcolor{arcol!81.7}81.7& \cellcolor{avgcol!74.8}74.8 & \cellcolor{timecol!05.56}$\phantom{0}$0.56 \\ 
        12 & Coupled Iterative Refinement (CIR) \cite{lipson2022coupled}  & \cellcolor{arcol!73.4}73.4	&\cellcolor{arcol!77.6}77.6	&\cellcolor{arcol!96.8}96.8	&\cellcolor{arcol!67.6}67.6&	\cellcolor{arcol!38.1}38.1&	\cellcolor{arcol!75.7}75.7&	\cellcolor{arcol!89.3}89.3& \cellcolor{avgcol!74.1}74.1& $\phantom{0}$-- \\ 
        13 & GDRNPP-PBRReal-RGB-MModel\cite{Wang_2021_GDRN,liu2022gdrnpp_bop} & \cellcolor{arcol!71.3}71.3	&\cellcolor{arcol!78.6}78.6	&\cellcolor{arcol!83.1}83.1	&\cellcolor{arcol!62.3}62.3&	\cellcolor{arcol!44.8}44.8&	\cellcolor{arcol!86.9}86.9&	\cellcolor{arcol!82.5}82.5& \cellcolor{avgcol!72.8}72.8 & \cellcolor{timecol!02.29}$\phantom{0}$0.23 \\ 
        14 & ZebraPoseSAT-EffnetB4~\cite{su2022zebrapose} & \cellcolor{arcol!72.1}72.1	&\cellcolor{arcol!80.6}80.6	&\cellcolor{arcol!85.0}85.0	&\cellcolor{arcol!54.5}54.5&	\cellcolor{arcol!41.0}41.0&	\cellcolor{arcol!88.2}88.2&	\cellcolor{arcol!83.0}83.0& \cellcolor{avgcol!72.0}72.0 & \cellcolor{timecol!02.50}$\phantom{0}$0.25 \\ 
        15 & ZebraPoseSAT-EffnetB4(DefaultDet)~\cite{su2022zebrapose}  & \cellcolor{arcol!70.7}70.7	&\cellcolor{arcol!76.8}76.8	&\cellcolor{arcol!84.9}84.9	&\cellcolor{arcol!59.7}59.7&	\cellcolor{arcol!41.7}41.7&	\cellcolor{arcol!88.7}88.7&	\cellcolor{arcol!81.6}81.6& \cellcolor{avgcol!72.0}72.0 & \cellcolor{timecol!02.50}$\phantom{0}$0.25 \\ 
        16 & ZebraPose-SAT~\cite{su2022zebrapose} & \cellcolor{arcol!72.1}72.1	&\cellcolor{arcol!78.7}78.7	&\cellcolor{arcol!86.1}86.1	&\cellcolor{arcol!54.9}54.9&	\cellcolor{arcol!37.9}37.9&	\cellcolor{arcol!84.7}84.7&	\cellcolor{arcol!82.8}82.8& \cellcolor{avgcol!71.0}71.0& $\phantom{0}$-- \\ 
        17 & Extended\_FCOS+PFA-MixPBR-RGB~\cite{hu2022perspective} & \cellcolor{arcol!74.5}74.5	&\cellcolor{arcol!77.8}77.8	&\cellcolor{arcol!83.9}83.9	&\cellcolor{arcol!60.0}60.0&	\cellcolor{arcol!35.3}35.3&	\cellcolor{arcol!84.1}84.1&	\cellcolor{arcol!80.6}80.6& \cellcolor{avgcol!70.9}70.9 & \cellcolor{timecol!30.19}$\phantom{0}$3.02 \\ 
        18 & GDRNPP-PBR-RGB-MModel \cite{Wang_2021_GDRN,liu2022gdrnpp_bop} & \cellcolor{arcol!71.3}71.3	&\cellcolor{arcol!79.6}79.6	&\cellcolor{arcol!75.2}75.2	&\cellcolor{arcol!62.3}62.3&	\cellcolor{arcol!44.8}44.8&	\cellcolor{arcol!86.9}86.9&	\cellcolor{arcol!71.3}71.3& \cellcolor{avgcol!70.2}70.2 & \cellcolor{timecol!02.84}$\phantom{0}$0.28 \\ 
        19 & CosyPose-ECCV20-SYNT+REAL-ICP~\cite{labbe2020cosypose} & \cellcolor{arcol!71.4}71.4	&\cellcolor{arcol!70.1}70.1	&\cellcolor{arcol!93.9}93.9	&\cellcolor{arcol!64.7}64.7&	\cellcolor{arcol!31.3}31.3&	\cellcolor{arcol!71.2}71.2&	\cellcolor{arcol!86.1}86.1& \cellcolor{avgcol!69.8}69.8 & \cellcolor{timecol!100}13.74 \\ 
        20 & ZebraPoseSAT-EffnetB4 (PBR\_Only)~\cite{su2022zebrapose} & \cellcolor{arcol!72.1}72.1	&\cellcolor{arcol!72.3}72.3	&\cellcolor{arcol!71.7}71.7	&\cellcolor{arcol!54.5}54.5&	\cellcolor{arcol!41.0}41.0&	\cellcolor{arcol!88.2}88.2&	\cellcolor{arcol!69.1}69.1& \cellcolor{avgcol!67.0}67.0& $\phantom{0}$-- \\ 
        21 & PFA-cosypose~\cite{hu2022perspective,labbe2020cosypose} & \cellcolor{arcol!71.4}71.4	&\cellcolor{arcol!73.8}73.8	&\cellcolor{arcol!83.7}83.7	&\cellcolor{arcol!59.6}59.6&	\cellcolor{arcol!24.6}24.6&	\cellcolor{arcol!71.2}71.2&	\cellcolor{arcol!80.7}80.7& \cellcolor{avgcol!66.4}66.4& $\phantom{0}$-- \\ 
        22 & Extended\_FCOS+PFA-PBR-RGB~\cite{hu2022perspective} & \cellcolor{arcol!74.5}74.5	&\cellcolor{arcol!71.9}71.9	&\cellcolor{arcol!73.2}73.2	&\cellcolor{arcol!60.0}60.0&	\cellcolor{arcol!35.3}35.3&	\cellcolor{arcol!84.1}84.1&	\cellcolor{arcol!64.8}64.8& \cellcolor{avgcol!66.3}66.3 & \cellcolor{timecol!34.97}$\phantom{0}$3.50 \\ 
        23 & SurfEmb-PBR-RGB~\cite{haugaard2022surfemb} & \cellcolor{arcol!66.3}66.3	&\cellcolor{arcol!73.5}73.5	&\cellcolor{arcol!71.5}71.5	&\cellcolor{arcol!58.8}58.8&	\cellcolor{arcol!41.3}41.3&	\cellcolor{arcol!79.1}79.1&	\cellcolor{arcol!64.7}64.7& \cellcolor{avgcol!65.0}65.0 & \cellcolor{timecol!88.91}$\phantom{0}$8.89 \\ 
        24 & Koenig-Hybrid-DL-PointPairs~\cite{koenig2020hybrid} & \cellcolor{arcol!63.1}63.1	&\cellcolor{arcol!65.5}65.5	&\cellcolor{arcol!92.0}92.0	&\cellcolor{arcol!43.0}43.0&	\cellcolor{arcol!48.3}48.3&	\cellcolor{arcol!65.1}65.1&	\cellcolor{arcol!70.1}70.1& \cellcolor{avgcol!63.9}63.9 & \cellcolor{timecol!06.33}$\phantom{0}$0.63 \\ 
        25 & CosyPose-ECCV20-SYNT+REAL-1VIEW~\cite{labbe2020cosypose}  & \cellcolor{arcol!63.3}63.3	&\cellcolor{arcol!72.8}72.8	&\cellcolor{arcol!82.3}82.3	&\cellcolor{arcol!58.3}58.3&	\cellcolor{arcol!21.6}21.6&	\cellcolor{arcol!65.6}65.6&	\cellcolor{arcol!82.1}82.1& \cellcolor{avgcol!63.7}63.7 & \cellcolor{timecol!04.49}$\phantom{0}$0.45 \\ 
        26 & CRT-6D & \cellcolor{arcol!66.0}66.0	&\cellcolor{arcol!64.4}64.4	&\cellcolor{arcol!78.9}78.9	&\cellcolor{arcol!53.7}53.7&	\cellcolor{arcol!20.8}20.8&	\cellcolor{arcol!60.3}60.3&	\cellcolor{arcol!75.2}75.2& \cellcolor{avgcol!59.9}59.9 & \cellcolor{timecol!0.059}$\phantom{0}$0.06 \\ 
        27 & Pix2Pose-BOP20\_w/ICP-ICCV19~\cite{park2019pix2pose} & \cellcolor{arcol!58.8}58.8	&\cellcolor{arcol!51.2}51.2	&\cellcolor{arcol!82.0}82.0	&\cellcolor{arcol!39.0}39.0&	\cellcolor{arcol!35.1}35.1&	\cellcolor{arcol!69.5}69.5&	\cellcolor{arcol!78.0}78.0& \cellcolor{avgcol!59.1}59.1 & \cellcolor{timecol!48.44}$\phantom{0}$4.84 \\ 
        28 & ZTE\_PPF & \cellcolor{arcol!66.3}66.3	&\cellcolor{arcol!37.4}37.4	&\cellcolor{arcol!90.4}90.4	&\cellcolor{arcol!39.6}39.6&	\cellcolor{arcol!47.0}47.0&	\cellcolor{arcol!73.5}73.5&	\cellcolor{arcol!50.2}50.2& \cellcolor{avgcol!57.8}57.8 & \cellcolor{timecol!0.901}$\phantom{0}$0.90 \\ 
        29 & CosyPose-ECCV20-PBR-1VIEW~\cite{labbe2020cosypose}  & \cellcolor{arcol!63.3}63.3	&\cellcolor{arcol!64.0}64.0	&\cellcolor{arcol!68.5}68.5	&\cellcolor{arcol!58.3}58.3&	\cellcolor{arcol!21.6}21.6&	\cellcolor{arcol!65.6}65.6&	\cellcolor{arcol!57.4}57.4& \cellcolor{avgcol!57.0}57.0 & \cellcolor{timecol!04.75}$\phantom{0}$0.48 \\ 
        30 & Vidal-Sensors18~\cite{vidal2018method}& \cellcolor{arcol!58.2}58.2	&\cellcolor{arcol!53.8}53.8	&\cellcolor{arcol!87.6}87.6	&\cellcolor{arcol!39.3}39.3&	\cellcolor{arcol!43.5}43.5&	\cellcolor{arcol!70.6}70.6&	\cellcolor{arcol!45.0}45.0& \cellcolor{avgcol!56.9}56.9 & \cellcolor{timecol!32.20}$\phantom{0}$3.22 \\ 
        31 & CDPNv2\_BOP20 (RGB-only \& ICP)~\cite{li2019cdpn} & \cellcolor{arcol!63.0}63.0	&\cellcolor{arcol!46.4}46.4	&\cellcolor{arcol!91.3}91.3	&\cellcolor{arcol!45.0}45.0&	\cellcolor{arcol!18.6}18.6&	\cellcolor{arcol!71.2}71.2&	\cellcolor{arcol!61.9}61.9& \cellcolor{avgcol!56.8}56.8 & \cellcolor{timecol!14.62}$\phantom{0}$1.46 \\ 
        32 & Drost-CVPR10-Edges~\cite{drost2010model} & \cellcolor{arcol!51.5}51.5	&\cellcolor{arcol!50.0}50.0	&\cellcolor{arcol!85.1}85.1	&\cellcolor{arcol!36.8}36.8&	\cellcolor{arcol!57.0}57.0&	\cellcolor{arcol!67.1}67.1&	\cellcolor{arcol!37.5}37.5& \cellcolor{avgcol!55.0}55.0 & \cellcolor{timecol!100}87.57 \\ 
        33 & CDPNv2\_BOP20 (PBR-only \& ICP)~\cite{li2019cdpn}  & \cellcolor{arcol!63.0}63.0	&\cellcolor{arcol!43.5}43.5	&\cellcolor{arcol!79.1}79.1	&\cellcolor{arcol!45.0}45.0&	\cellcolor{arcol!18.6}18.6&	\cellcolor{arcol!71.2}71.2&	\cellcolor{arcol!53.2}53.2& \cellcolor{avgcol!53.4}53.4 & \cellcolor{timecol!14.91}$\phantom{0}$1.49 \\ 
        34 & CDPNv2\_BOP20 (RGB-only)~\cite{li2019cdpn}  & \cellcolor{arcol!62.4}62.4	&\cellcolor{arcol!47.8}47.8	&\cellcolor{arcol!77.2}77.2	&\cellcolor{arcol!47.3}47.3&	\cellcolor{arcol!10.2}10.2&	\cellcolor{arcol!72.2}72.2&	\cellcolor{arcol!53.2}53.2& \cellcolor{avgcol!52.9}52.9 & \cellcolor{timecol!09.35}$\phantom{0}$0.94 \\ 
        35 & Drost-CVPR10-3D-Edges~\cite{drost2010model}& \cellcolor{arcol!46.9}46.9	&\cellcolor{arcol!40.4}40.4	&\cellcolor{arcol!85.2}85.2	&\cellcolor{arcol!37.3}37.3&	\cellcolor{arcol!46.2}46.2&	\cellcolor{arcol!62.3}62.3&	\cellcolor{arcol!31.6}31.6& \cellcolor{avgcol!50.0}50.0 & \cellcolor{timecol!100}80.06 \\ 
        36 & Drost-CVPR10-3D-Only~\cite{drost2010model}& \cellcolor{arcol!52.7}52.7	&\cellcolor{arcol!44.4}44.4	&\cellcolor{arcol!77.5}77.5	&\cellcolor{arcol!38.8}38.8&	\cellcolor{arcol!31.6}31.6&	\cellcolor{arcol!61.5}61.5&	\cellcolor{arcol!34.4}34.4& \cellcolor{avgcol!48.7}48.7 & \cellcolor{timecol!77.04}$\phantom{0}$7.70 \\ 
        37 & CDPN\_BOP19 (RGB-only)~\cite{li2019cdpn}  & \cellcolor{arcol!56.9}56.9	&\cellcolor{arcol!49.0}49.0	&\cellcolor{arcol!76.9}76.9	&\cellcolor{arcol!32.7}32.7 & $\phantom{0}$\cellcolor{arcol!6.7}6.7&	\cellcolor{arcol!67.2}67.2&	\cellcolor{arcol!45.7}45.7& \cellcolor{avgcol!47.9}47.9 & \cellcolor{timecol!04.80}$\phantom{0}$0.48 \\ 
        38 & CDPNv2\_BOP20 (PBR-only \& RGB-only)~\cite{li2019cdpn} & \cellcolor{arcol!62.4}62.4	&\cellcolor{arcol!40.7}40.7	&\cellcolor{arcol!58.8}58.8	&\cellcolor{arcol!47.3}47.3&	\cellcolor{arcol!10.2}10.2&	\cellcolor{arcol!72.2}72.2&	\cellcolor{arcol!39.0}39.0& \cellcolor{avgcol!47.2}47.2 & \cellcolor{timecol!09.78}$\phantom{0}$0.98 \\ 
        39 & leaping from 2D to 6D~\cite{liu2010leaping} & \cellcolor{arcol!52.5}52.5	&\cellcolor{arcol!40.3}40.3	&\cellcolor{arcol!75.1}75.1	&\cellcolor{arcol!34.2}34.2&	$\phantom{0}$\cellcolor{arcol!7.7}7.7&	\cellcolor{arcol!65.8}65.8&	\cellcolor{arcol!54.3}54.3& \cellcolor{avgcol!47.1}47.1 & \cellcolor{timecol!04.25}$\phantom{0}$0.43 \\ 
        40 & EPOS-BOP20-PBR ~\cite{hodan2020epos} & \cellcolor{arcol!54.7}54.7	&\cellcolor{arcol!46.7}46.7	&\cellcolor{arcol!55.8}55.8	&\cellcolor{arcol!36.3}36.3&	\cellcolor{arcol!18.6}18.6&	\cellcolor{arcol!58.0}58.0&	\cellcolor{arcol!49.9}49.9& \cellcolor{avgcol!45.7}45.7 & \cellcolor{timecol!18.74}$\phantom{0}$1.87 \\ 
        41 & Drost-CVPR10-3D-Only-Faster~\cite{drost2010model}& \cellcolor{arcol!49.2}49.2	&\cellcolor{arcol!40.5}40.5	&\cellcolor{arcol!69.6}69.6	&\cellcolor{arcol!37.7}37.7&	\cellcolor{arcol!27.4}27.4&	\cellcolor{arcol!60.3}60.3&	\cellcolor{arcol!33.0}33.0& \cellcolor{avgcol!45.4}45.4 & \cellcolor{timecol!13.83}$\phantom{0}$1.38 \\ 
        42 & Félix\&Neves-ICRA2017-IET2019~\cite{rodrigues2019deep,raposo2017using}& \cellcolor{arcol!39.4}39.4	&\cellcolor{arcol!21.2}21.2	&\cellcolor{arcol!85.1}85.1	&\cellcolor{arcol!32.3}32.3&	$\phantom{0}$\cellcolor{arcol!6.9}6.9&	\cellcolor{arcol!52.9}52.9&	\cellcolor{arcol!51.0}51.0& \cellcolor{avgcol!41.2}41.2 & \cellcolor{timecol!100}55.78 \\ 
        43 & Sundermeyer-IJCV19+ICP~\cite{sundermeyer2019augmented}  & \cellcolor{arcol!23.7}23.7	&\cellcolor{arcol!48.7}48.7	&\cellcolor{arcol!61.4}61.4	&\cellcolor{arcol!28.1}28.1&	\cellcolor{arcol!15.8}15.8&	\cellcolor{arcol!50.6}50.6&	\cellcolor{arcol!50.5}50.5& \cellcolor{avgcol!39.8}39.8 & \cellcolor{timecol!08.65}$\phantom{0}$0.86 \\ 
        44 & Zhigang-CDPN-ICCV19~\cite{li2019cdpn} & \cellcolor{arcol!37.4}37.4	&\cellcolor{arcol!12.4}12.4	&\cellcolor{arcol!75.7}75.7	&\cellcolor{arcol!25.7}25.7&	$\phantom{0}$\cellcolor{arcol!7.0}7.0&	\cellcolor{arcol!47.0}47.0&	\cellcolor{arcol!42.2}42.2& \cellcolor{avgcol!35.3}35.3 & \cellcolor{timecol!05.13}$\phantom{0}$0.51 \\ 
        45 & PointVoteNet2~\cite{hagelskjaer2019pointposenet} & \cellcolor{arcol!65.3}65.3	& $\phantom{0}$\cellcolor{arcol!0.4}0.4	&\cellcolor{arcol!67.3}67.3	& \cellcolor{arcol!26.4}26.4&	$\phantom{0}$\cellcolor{arcol!0.1}0.1&	\cellcolor{arcol!55.6}55.6&	\cellcolor{arcol!30.8}30.8& \cellcolor{avgcol!35.1}35.1& $\phantom{0}$-- \\ 
        46 & Pix2Pose-BOP20-ICCV19 ~\cite{park2019pix2pose}& \cellcolor{arcol!36.3}36.3	&\cellcolor{arcol!34.4}34.4	&\cellcolor{arcol!42.0}42.0	&\cellcolor{arcol!22.6}22.6&	\cellcolor{arcol!13.4}13.4&	\cellcolor{arcol!44.6}44.6&	\cellcolor{arcol!45.7}45.7& \cellcolor{avgcol!34.2}34.2 & \cellcolor{timecol!12.15}$\phantom{0}$1.22 \\ 
        47 & Sundermeyer-IJCV19\cite{sundermeyer2019augmented} & \cellcolor{arcol!14.6}14.6	&\cellcolor{arcol!30.4}30.4	&\cellcolor{arcol!40.1}40.1	&\cellcolor{arcol!21.7}21.7&	\cellcolor{arcol!10.1}10.1&	\cellcolor{arcol!34.6}34.6&	\cellcolor{arcol!44.6}44.6& \cellcolor{avgcol!28.0}28.0 & \cellcolor{timecol!01.96}$\phantom{0}$0.20 \\ 
        48 & SingleMultiPathEncoder-CVPR20 \cite{sundermeyer2020multi} & \cellcolor{arcol!21.7}21.7	&\cellcolor{arcol!31.0}31.0	&\cellcolor{arcol!33.4}33.4	&\cellcolor{arcol!17.5}17.5&	$\phantom{0}$\cellcolor{arcol!6.7}6.7&	\cellcolor{arcol!29.3}29.3&	\cellcolor{arcol!28.9}28.9& \cellcolor{avgcol!24.1}24.1 & \cellcolor{timecol!01.86}$\phantom{0}$0.19 \\ 
        49 & DPOD (synthetic) ~\cite{zakharov2019dpod} & \cellcolor{arcol!16.9}16.9	& $\phantom{0}$\cellcolor{arcol!8.1}8.1	&\cellcolor{arcol!24.2}24.2	&\cellcolor{arcol!13.0}13.0&	$\phantom{0}$\cellcolor{arcol!0.0}0.0&	\cellcolor{arcol!28.6}28.6&	\cellcolor{arcol!22.2}22.2& \cellcolor{avgcol!16.1}16.1 & \cellcolor{timecol!02.31}$\phantom{0}$0.23 \\
        \bottomrule
  
    \end{tabularx}
    \caption{\textbf{6D object localization results on the seven core datasets.}
    The methods are ranked by the $\text{AR}_C$ score which is the average of the per-dataset $\text{AR}_D$ scores defined in Sec.~\ref{sec:6d_localization_task}. The last column shows the average image processing time (in seconds).
    }
    \label{tab:pose_methods_results}

    \vspace{0.5ex}
\end{table*}

\setlength{\tabcolsep}{2pt}
\begin{table*}[t]
    \footnotesize
    \begin{tabularx}{\linewidth}{rLllllllll}
        \toprule
        \# & Method & Year & Type & DNN per & Det./seg. & Refinement & Train im. & ...type & Test im. \\ %
        \midrule
        1 & GDRNPP-PBRReal-RGBD-MModel \cite{Wang_2021_GDRN,liu2022gdrnpp_bop}& \cellcolor{ccol!100}2022 & DNN & Object & YOLOX & \cellcolor{ccol!100}${\tiny\sim}$CIR & RGB-D & PBR+real & RGB-D \\ %
        2 & GDRNPP-PBR-RGBD-MModel \cite{Wang_2021_GDRN,liu2022gdrnpp_bop}& \cellcolor{ccol!100}2022 & DNN & Object & YOLOX & \cellcolor{ccol!100}${\tiny\sim}$CIR & RGB-D & \cellcolor{ccol!100}PBR & RGB-D \\ %
        3 & GDRNPP-PBRReal-RGBD-MModel-Fast \cite{Wang_2021_GDRN,liu2022gdrnpp_bop}& \cellcolor{ccol!100}2022 & DNN & Object & YOLOX & \cellcolor{ccol!100}Depth adjust. & \cellcolor{ccol!100}RGB & PBR+real & RGB-D \\ %
        4 & GDRNPP-PBRReal-RGBD-MModel-Offi. \cite{Wang_2021_GDRN,liu2022gdrnpp_bop}& \cellcolor{ccol!100}2022 & DNN & Object & \cellcolor{ccol!100}Default (synt+real) & \cellcolor{ccol!100}${\tiny\sim}$CIR & RGB-D & PBR+real & RGB-D \\ %
        5 & Extended\_FCOS+PFA-MixPBR-RGBD~\cite{hu2022perspective} & \cellcolor{ccol!100}2022 & DNN & \cellcolor{ccol!100}Dataset & Extended FCOS & \cellcolor{ccol!100}PFA & \cellcolor{ccol!100}RGB & PBR+real & RGB-D \\ %
        6 & Extended\_FCOS+PFA-MixPBR-RGBD-Fast~\cite{hu2022perspective} & \cellcolor{ccol!100}2022 & DNN & \cellcolor{ccol!100}Dataset & Extended FCOS & \cellcolor{ccol!100}PFA & \cellcolor{ccol!100}RGB & PBR+real & RGB-D \\ %
        7 & RCVPose3D-SingleModel-VIVO-PBR~\cite{wu2022keypoint} & \cellcolor{ccol!100}2022 & DNN & \cellcolor{ccol!100}Dataset & RCVPose3D & \cellcolor{ccol!100}ICP & RGB-D & PBR+real & RGB-D \\ %
        8 & ZebraPoseSAT-EffnetB4+ICP(DefaultDet)~\cite{su2022zebrapose} & \cellcolor{ccol!100}2022 & DNN & Object & \cellcolor{ccol!100}Default (synt+real) & \cellcolor{ccol!100}ICP & \cellcolor{ccol!100}RGB & PBR+real & RGB-D \\ %
        9 & Extended\_FCOS+PFA-PBR-RGBD~\cite{hu2022perspective} & \cellcolor{ccol!100}2022 & DNN & \cellcolor{ccol!100}Dataset & Extended FCOS & \cellcolor{ccol!100}PFA & \cellcolor{ccol!100}RGB & \cellcolor{ccol!100}PBR & RGB-D \\ %
        10 & SurfEmb-PBR-RGBD~\cite{haugaard2022surfemb} & \cellcolor{ccol!100}2022 & DNN & \cellcolor{ccol!100}Dataset & \cellcolor{ccol!100}Default (PBR) & \cellcolor{ccol!100}Custom & RGB-D & \cellcolor{ccol!100}PBR & RGB-D \\ %
        11 & GDRNPP-PBRReal-RGBD-SModel \cite{Wang_2021_GDRN,liu2022gdrnpp_bop} & \cellcolor{ccol!100}2022 & DNN & \cellcolor{ccol!100}Dataset & YOLOX & \cellcolor{ccol!100}Depth adjust. & \cellcolor{ccol!100}RGB & PBR+real & RGB-D \\ %
        12 & Coupled Iterative Refinement (CIR) \cite{lipson2022coupled} & \cellcolor{ccol!100}2022 & DNN & \cellcolor{ccol!100}Dataset & \cellcolor{ccol!100}Default (synt+real) & \cellcolor{ccol!100}CIR & RGB-D & PBR+real & RGB-D \\ %
        13 & GDRNPP-PBRReal-RGB-MModel\cite{Wang_2021_GDRN,liu2022gdrnpp_bop} & \cellcolor{ccol!100}2022 & DNN & Object & YOLOX & -- & \cellcolor{ccol!100}RGB & PBR+real & \cellcolor{ccol!100}RGB \\ %
        14 & ZebraPoseSAT-EffnetB4~\cite{su2022zebrapose} & \cellcolor{ccol!100}2022 & DNN & Object & FCOS & -- & \cellcolor{ccol!100}RGB & PBR+real & \cellcolor{ccol!100}RGB \\ %
        15 & ZebraPoseSAT-EffnetB4(DefaultDet)~\cite{su2022zebrapose} & \cellcolor{ccol!100}2022 & DNN & Object & \cellcolor{ccol!100}Default (synt+real) & -- & \cellcolor{ccol!100}RGB & PBR+real & \cellcolor{ccol!100}RGB \\ %
        16 & ZebraPose-SAT~\cite{su2022zebrapose} & \cellcolor{ccol!100}2022 & DNN & Object & FCOS & -- & \cellcolor{ccol!100}RGB & PBR+real & \cellcolor{ccol!100}RGB \\ %
        17 & Extended\_FCOS+PFA-MixPBR-RGB~\cite{hu2022perspective} & \cellcolor{ccol!100}2022 & DNN & \cellcolor{ccol!100}Dataset & Extended FCOS & \cellcolor{ccol!100}PFA & \cellcolor{ccol!100}RGB & PBR+real & \cellcolor{ccol!100}RGB \\ %
        18 & GDRNPP-PBR-RGB-MModel \cite{Wang_2021_GDRN,liu2022gdrnpp_bop} & \cellcolor{ccol!100}2022 & DNN & Object & YOLOX & -- & \cellcolor{ccol!100}RGB & \cellcolor{ccol!100}PBR & \cellcolor{ccol!100}RGB \\ %
        19 & CosyPose-ECCV20-SYNT+REAL-ICP~\cite{labbe2020cosypose} & 2020 & DNN & \cellcolor{ccol!100}Dataset & \cellcolor{ccol!100}Default (synt+real) & \cellcolor{ccol!100}DeepIM+ICP & \cellcolor{ccol!100}RGB & PBR+real & RGB-D \\ %
        20 & ZebraPoseSAT-EffnetB4 (PBR\_Only)~\cite{su2022zebrapose} & \cellcolor{ccol!100}2022 & DNN & Object & FCOS & -- & \cellcolor{ccol!100}RGB & \cellcolor{ccol!100}PBR & \cellcolor{ccol!100}RGB \\ %
        21 & PFA-cosypose~\cite{hu2022perspective,labbe2020cosypose} & \cellcolor{ccol!100}2022 & DNN & \cellcolor{ccol!100}Dataset & MaskRCNN & \cellcolor{ccol!100}PFA & RGB-D & PBR+real & \cellcolor{ccol!100}RGB \\ %
        22 & Extended\_FCOS+PFA-PBR-RGB~\cite{hu2022perspective} & \cellcolor{ccol!100}2022 & DNN & \cellcolor{ccol!100}Dataset & Extended FCOS & \cellcolor{ccol!100}PFA & \cellcolor{ccol!100}RGB & \cellcolor{ccol!100}PBR & \cellcolor{ccol!100}RGB \\ %
        23 & SurfEmb-PBR-RGB~\cite{haugaard2022surfemb} & \cellcolor{ccol!100}2022 & DNN & \cellcolor{ccol!100}Dataset & \cellcolor{ccol!100}Default (PBR) & \cellcolor{ccol!100}Custom & \cellcolor{ccol!100}RGB & \cellcolor{ccol!100}PBR & \cellcolor{ccol!100}RGB \\ %
        24 & Koenig-Hybrid-DL-PointPairs~\cite{koenig2020hybrid} & 2020 & DNN/PPF \cellcolor{ccol!100}& \cellcolor{ccol!100}Dataset & Retina/MaskRCNN & \cellcolor{ccol!100}{\raggedright ICP} & \cellcolor{ccol!100}RGB & Synt+real & RGB-D \\ %
        25 & CosyPose-ECCV20-SYNT+REAL-1VIEW~\cite{labbe2020cosypose} & 2020 & DNN & \cellcolor{ccol!100}Dataset & \cellcolor{ccol!100}Default (synt+real) & \cellcolor{ccol!100}${\tiny\sim}$DeepIM & \cellcolor{ccol!100}RGB & PBR+real & \cellcolor{ccol!100}RGB \\ %
        26 & CRT-6D & \cellcolor{ccol!100}2022 & DNN & \cellcolor{ccol!100}Dataset & \cellcolor{ccol!100}Default (synt+real) & \cellcolor{ccol!100}Custom & \cellcolor{ccol!100}RGB & PBR+real & \cellcolor{ccol!100}RGB \\ %
        27 & Pix2Pose-BOP20\_w/ICP-ICCV19~\cite{park2019pix2pose} & 2020 & DNN & Object & MaskRCNN & \cellcolor{ccol!100}ICP & \cellcolor{ccol!100}RGB & PBR+real & RGB-D \\ %
        28 & ZTE\_PPF & \cellcolor{ccol!100}2022 & DNN/PPF \cellcolor{ccol!100}& \cellcolor{ccol!100}Dataset & \cellcolor{ccol!100}Default (synt+real) & \cellcolor{ccol!100}ICP & \cellcolor{ccol!100}RGB & PBR+real & RGB-D \\ %
        29 & CosyPose-ECCV20-PBR-1VIEW~\cite{labbe2020cosypose} & 2020 & DNN & \cellcolor{ccol!100}Dataset & \cellcolor{ccol!100}Default (PBR) & \cellcolor{ccol!100}${\tiny\sim}$DeepIM & \cellcolor{ccol!100}RGB & \cellcolor{ccol!100}PBR & \cellcolor{ccol!100}RGB \\ %
        30 & Vidal-Sensors18~\cite{vidal2018method}  & 2019 & PPF \cellcolor{ccol!100}& -- & -- & \cellcolor{ccol!100}ICP & -- & -- & D \\ %
        31 & CDPNv2\_BOP20 (RGB-only \& ICP)~\cite{li2019cdpn} & 2020 & DNN & Object & FCOS & \cellcolor{ccol!100}ICP & \cellcolor{ccol!100}RGB & Synt+real & RGB-D \\ %
        32 & Drost-CVPR10-Edges~\cite{drost2010model} & 2019 & PPF \cellcolor{ccol!100}& -- & -- & \cellcolor{ccol!100}ICP & -- & -- & RGB-D \\ %
        33 & CDPNv2\_BOP20 (PBR-only \& ICP)~\cite{li2019cdpn}  & 2020 & DNN & Object & FCOS & \cellcolor{ccol!100}ICP & \cellcolor{ccol!100}RGB & \cellcolor{ccol!100}PBR & RGB-D \\ %
        34 & CDPNv2\_BOP20 (RGB-only)~\cite{li2019cdpn}  & 2020 & DNN & Object & FCOS & -- & \cellcolor{ccol!100}RGB & Synt+real & \cellcolor{ccol!100}RGB \\ %
        35 & Drost-CVPR10-3D-Edges~\cite{drost2010model} & 2019 & PPF \cellcolor{ccol!100}& -- & -- & \cellcolor{ccol!100}ICP & -- & -- & D \\ %
        36 & Drost-CVPR10-3D-Only~\cite{drost2010model} & 2019 & PPF \cellcolor{ccol!100}& -- & -- & \cellcolor{ccol!100}ICP & -- & -- & D \\ %
        37 & CDPN\_BOP19 (RGB-only)~\cite{li2019cdpn} & 2020 & DNN & Object & RetinaNet & -- & \cellcolor{ccol!100}RGB & Synt+real & \cellcolor{ccol!100}RGB \\ %
        38 & CDPNv2\_BOP20 (PBR-only \& RGB-only)~\cite{li2019cdpn} & 2020 & DNN & Object & FCOS & -- & \cellcolor{ccol!100}RGB & \cellcolor{ccol!100}PBR & \cellcolor{ccol!100}RGB \\ %
        39 & leaping from 2D to 6D~\cite{liu2010leaping} & 2020 & DNN & Object & Unknown & -- & \cellcolor{ccol!100}RGB & Synt+real & \cellcolor{ccol!100}RGB \\ %
        40 & EPOS-BOP20-PBR ~\cite{hodan2020epos} & 2020 & DNN & \cellcolor{ccol!100}Dataset & -- & -- & \cellcolor{ccol!100}RGB & \cellcolor{ccol!100}PBR & \cellcolor{ccol!100}RGB \\ %
        41 & Drost-CVPR10-3D-Only-Faster~\cite{drost2010model} & 2019 & PPF \cellcolor{ccol!100}& -- & -- & \cellcolor{ccol!100}ICP & -- & -- & D \\ %
        42 & Félix\&Neves-ICRA2017-IET2019~\cite{rodrigues2019deep,raposo2017using} & 2019 & DNN/PPF \cellcolor{ccol!100}& \cellcolor{ccol!100}Dataset & MaskRCNN & \cellcolor{ccol!100}ICP & RGB-D & Synt+real & RGB-D \\ %
        43 & Sundermeyer-IJCV19+ICP~\cite{sundermeyer2019augmented} & 2019 & DNN & Object & RetinaNet & \cellcolor{ccol!100}ICP & \cellcolor{ccol!100}RGB & Synt+real & RGB-D \\ %
        44 & Zhigang-CDPN-ICCV19~\cite{li2019cdpn} & 2019 & DNN & Object & RetinaNet & -- & \cellcolor{ccol!100}RGB & Synt+real & \cellcolor{ccol!100}RGB \\ %
        45 & PointVoteNet2~\cite{hagelskjaer2019pointposenet} & 2020 & DNN & Object & -- & \cellcolor{ccol!100}ICP & RGB-D & \cellcolor{ccol!100}PBR & RGB-D \\ %
        46 & Pix2Pose-BOP20-ICCV19 ~\cite{park2019pix2pose}& 2020 & DNN & Object & MaskRCNN & -- & \cellcolor{ccol!100}RGB & PBR+real & \cellcolor{ccol!100}RGB \\ %
        47 & Sundermeyer-IJCV19\cite{sundermeyer2019augmented} & 2019 & DNN & Object & RetinaNet & -- & \cellcolor{ccol!100}RGB & Synt+real & \cellcolor{ccol!100}RGB \\ %
        48 & SingleMultiPathEncoder-CVPR20 \cite{sundermeyer2020multi}& 2020 & DNN & \cellcolor{ccol!100}All & MaskRCNN & -- & \cellcolor{ccol!100}RGB & Synt+real & \cellcolor{ccol!100}RGB \\ %
        49 & DPOD (synthetic) ~\cite{zakharov2019dpod} & 2019 & DNN & \cellcolor{ccol!100}Dataset & -- & -- & \cellcolor{ccol!100}RGB & Synt & \cellcolor{ccol!100}RGB \\ %
        \bottomrule
  
    \end{tabularx}
    \caption{\textbf{Properties of evaluated 6D object localization methods.} Column \emph{Year} is the year of submission, \emph{Type} indicates whether the method relies on deep neural networks (DNN's) or point pair features (PPF's), \emph{DNN per...}~shows how many DNN models were trained, \emph{Det./seg.}~is the object detection or segmentation method, \emph{Refinement} is the pose refinement method, \emph{Train im.}~and \emph{Test im.}~show image channels used at training and test time respectively, and \emph{Train im.~type} is the domain of training images. All test images are real.
    }
    \label{tab:pose_methods_properties}
    \vspace{0.5ex}
\end{table*}

\subsection{6D Object Localization Results} \label{sec:6d_results}

An overview of the 6D object localization results is in Tab.~\ref{tab:pose_methods_results} and properties of the evaluated methods in Tab.~\ref{tab:pose_methods_properties}.
In 2022, all 23 of the new submissions rely on DNN's in their pipelines and 18 of them outperform CosyPose~\cite{labbe2020cosypose}, the top-performing method from the 2020 challenge.
The best method from 2022, GDRNPP~\cite{Wang_2021_GDRN, liu2022gdrnpp_bop}, is purely learning-based and achieves 83.7 AR$_C$, outperforming CosyPose by substantial 13.9 points in AR$_C$ (\#1$-$\#19 in Tab.~\ref{tab:pose_methods_results}).
Gains in accuracy are most notable on the industrial ITODD dataset~\cite{drost2017introducing} where GDRNPP reaches 67.9 AR$_C$ (+36.6 AR$_C$ \wrt CosyPose). This result is significant as ITODD reflects a challenging industrial scenario and was previously dominated by PPF-based approaches, the best of which, KoenigHybrid~\cite{koenig2020hybrid} (\#24), achieved 48.3 AR$_C$.

\customparagraph{GDRNPP dominates in 2022:} 
The GDRNPP method was evaluated in seven variants, four of which are on top of the leaderboard.
The variants were tailored towards different BOP 2022 awards (Sec.~\ref{sec:awards}) by relying on different data domains and modalities and on different detection and pose refinement methods. Having results of these variants enables to understand the importance of individual aspects of the pipeline.
The common ground is the Geometrically-Guided Direct Regression Network (GDR-Net)~\cite{Wang_2021_GDRN}, which takes an RGB object crop as input and densely predicts 2D-3D correspondences, identities of surface fragments~\cite{hodan2020epos}, and a mask of the visible object part. 
Then, instead of applying P\emph{n}P-RANSAC~\cite{hodan2020epos}, the predictions are concatenated and fed into a small CNN with a fully connected head that regresses a scale-invariant translation~\cite{li2019cdpn} and a 3D rotation using the allocentric 6D representation~\cite{kundu20183d}. The 3D rotation loss takes into account object symmetries that are provided in the BOP datasets.
For BOP 2022, GDR-Net~\cite{Wang_2021_GDRN} was modified by exchanging the ResNet34 backbone with ConvNext~\cite{Liu_2022_CVPR}, predicting both modal and amodal masks as intermediate representations, and applying stronger domain randomization.
The winning GDRNPP variant trains YOLOX~\cite{ge2021yolox} for object detection and GDR-Net for pose estimation on the provided PBR and real RGB images, and refines the poses by a multi-hypotheses refinement method inspired by Coupled Iterative Refinement (CIR)~\cite{lipson2022coupled}, which is trained on PBR and real RGB-D images.

\customparagraph{Training on depth:} Methods RCVPose3D~\cite{wu2022keypoint} (\#7) and CIR~\cite{lipson2022coupled} (\#12; a variant is also used in \#1, 2, 4), started benefiting from learning on the depth channel in addition to the RGB channels (only PointVoteNet2~\cite{hagelskjaer2019pointposenet} applied a neural network to the depth channel in 2020).
On the flip side, the multi-hypotheses refinement methods can be time-intensive -- the CIR-based approach increases the inference time of GDRNPP by 6.03s per image on average (\#1$-$\#3).

\customparagraph{Increased accuracy \& speed:} The third GDRNPP entry replaces the CIR-based refinement~\cite{lipson2022coupled}, which is used in the top two entries, by a fast and simple depth-based adjustment of the 3D translation and still achieves impressive 80.5 AR$_C$ in just 0.23s per image. In comparison, the best method in 2020 that took less than 1s per image is KoenigHybrid~\cite{koenig2020hybrid} (\#24) with 63.9 AR$_C$ and 0.63s per image.

\customparagraph{RGB-only from 2022 beats RGB-D from 2020:} The best method that relies only on RGB image channels at both training and test time is a variant of GDRNPP (\#13). Without any pose refinement, this method achieves 72.8 AR$_C$ which is +9.1 \wrt CosyPose that applies RGB-based pose refinement (\#25) and +3.0 \wrt to the overall best method from 2020, \ie, CosyPose with a
depth-based ICP (\#19).

\customparagraph{Synthetic-to-real gap shrinks further:} Another important result was achieved by the GDRNPP variant that is trained only on the provided synthetic PBR images rendered with BlenderProc~\cite{denninger2019blenderproc, denninger2020blenderproc}. With 82.7~AR$_C$, this variant achieves the second highest accuracy. On datasets with real training images (T-LESS, YCB-V, \mbox{TUD-L}), the synthetically trained variant is only -2.5 AR$_C$ on average behind the winning method that was trained on both PBR and real training images. In the RGB-only setting, the synthetic-to-real gap has been reduced on the three datasets from $\Delta$15.8 AR$_C$ (observed on CosyPose in 2020; \#25$-$\#29) to $\Delta$6.2 AR$_C$ (observed on GDRNPP in 2022; \#13$-$\#18).
The BOP 2020 results~\cite{hodan2020bop} demonstrated the importance of training on PBR images over training on rasterized images with random backgrounds.
The BOP 2022 results confirm this observation and also suggest that the synthetic-to-real gap monotonically shrinks as the accuracy of methods increases (see, \eg, \#25$-$\#29, \#14$-$\#20, \#5$-$\#9, \#1$-$\#2 in Tab.~\ref{tab:pose_methods_results}).

\customparagraph{Scalability in the number of objects:} The advancement in the synthetic-to-real transfer is crucial for increasing the scope of applications. In addition, real world applications require methods whose computational and memory resources scale gracefully with the amount of target objects. The top four GDRNPP variants are all trained with at least one pose network per object. This means that the training and inference time complexity and the inference memory increase linearly with the number of target objects. When GDRNPP is trained with one pose network per BOP dataset containing 2--33 objects (Tab.~\ref{tab:dataset_params}), it achieves only 74.8~AR$_C$ (\#11) and is outperformed by, \eg, Extended\_FCOS+PFA~\cite{hu2022perspective} (\#5) that reaches 78.7~AR$_C$ with one pose network per dataset. This raises the question how the results would change if \cite{hu2022perspective} was trained per object.

\customparagraph{2D detection followed by 6D pose estimation:} Almost all 6D object localization methods evaluated in 2022 start by detecting the object instances in RGB images by predicting their 2D bounding boxes. 
Some methods also predict 2D object masks in the detected regions at training time for loss calculation~\cite{hu2022perspective} or extra supervision~\cite{he2017mask}, and some predict 2D masks at both training and inference time and use them to establish correspondences~\cite{su2022zebrapose,haugaard2022surfemb}. 
The only exception is RCVPose3D~\cite{wu2022keypoint}, which does not start by detecting object instance in the RGB image channels and instead segments the object instances in 3D point clouds calculated from the depth image channel.

\customparagraph{Detector-agnostic results:}
Eleven methods use the default 2D object detections (\emph{Default} in column \emph{Det./seg.} in Tab.~\ref{tab:pose_methods_properties}), which were provided to participants of the 2022 challenge and produced by Mask R-CNN~\cite{he2017mask} trained for the first stage of CosyPose~\cite{labbe2020cosypose} in 2020. Three of these methods use detections from Mask R-CNN trained only on PBR images, and eight use detections from Mask R-CNN trained on synthetic and real images (where the synthetic include PBR and additional images synthesized by the authors of~\cite{labbe2020cosypose}).
Among the eight methods, GDRNPP is once again at the top with 79.8~AR$_C$ (\#4). We can therefore conclude that the pose estimation performance of the GDRNPP pipeline is performing best independent of the used detection method.
However, the accuracy gap to other methods decreases with the default detections, \eg, from +7.2~AR$_C$ (\#1$-$\#8) to +3.3~AR$_C$ (\#4$-$\#8) \wrt ZebraPose~\cite{su2022zebrapose}.

\subsection{2D Object Detection Results} \label{sec:det_results}

As shown in Tab.~\ref{tab:det_results}, the YOLOX~\cite{ge2021yolox} detector from GDRNPP has the top performance of 77.3~AP$_C$. This detector employs a ConvNext~\cite{Liu_2022_CVPR} backbone and was trained with the Ranger optimizer\cite{Ranger} and strong data augmentation.
Mask R-CNN~\cite{he2017mask} from CosyPose only achieves 60.5~AP$_C$ (-16.8~AP$_C$), which explains the +3.9~AR$_C$ gain in the pose accuracy (\#1$-$\#4 in Tab.~\ref{tab:pose_methods_results}).
YOLOX is relatively insensitive to the image domain, improving only +3.5 AP$_C$ (\#1$-$\#2 in Tab.~\ref{tab:det_results}) when trained also on real images. Mask R-CNN yields +4.8~AP$_C$ (\#6$-$\#7) and FCOS~\cite{tian2019fcos} yields +5.4~AP$_C$ (\#3$-$\#4) in such a comparison.

Although all 2D object detection methods rely only on RGB and ignore the depth channel, they work remarkably well even on the texture-less objects from T-LESS~\cite{hodan2017tless} (see the BOP website for per-dataset scores).
However, detections from YOLOX on YCB-V~\cite{xiang2017posecnn} in Fig.~\ref{fig:cover_results} reveal a limitation of the RGB-only detection that fails to distinguish the two differently sized clamps.
This detection failure can cause
wrong pose estimates even though the rendered scene seems perfectly plausible. Depth data could help to disambiguate the object scale in such cases.

\setlength{\tabcolsep}{2pt}
\begin{table}[t]
    \footnotesize
    \begin{tabularx}{\columnwidth}{lLllllll} \toprule
        \# & Method & ...based on & Year & Data & ...type & AP$_C$ & Time \\ 
        \toprule
        1 & GDRNPPDet & YOLOX & \cellcolor{ccol!100}2022 & \cellcolor{ccol!100}RGB & PBR+real &  \cellcolor{avgcol!77.3}77.3 & \cellcolor{timecol!41}.081 \\ 
        2 & GDRNPPDet & YOLOX & \cellcolor{ccol!100}2022 & \cellcolor{ccol!100}RGB & \cellcolor{ccol!100}PBR & \cellcolor{avgcol!73.8}73.8 & \cellcolor{timecol!41}.081 \\ 
        3 & Extended\_FCOS & FCOS & \cellcolor{ccol!100}2022 & \cellcolor{ccol!100}RGB & PBR+real & \cellcolor{avgcol!72.1}72.1 & \cellcolor{timecol!15}.030 \\ 
        4 & Extended\_FCOS & FCOS & \cellcolor{ccol!100}2022 & \cellcolor{ccol!100}RGB & \cellcolor{ccol!100}PBR & \cellcolor{avgcol!66.7}66.7 & \cellcolor{timecol!15}.030 \\ 
        5 & DLZDet & DLZDet & \cellcolor{ccol!100}2022 & \cellcolor{ccol!100}RGB & \cellcolor{ccol!100}PBR & \cellcolor{avgcol!65.6}65.6 & - \\ 
        6 & CosyPose & Mask R-CNN & 2020 & \cellcolor{ccol!100}RGB & PBR+real & \cellcolor{avgcol!60.5}60.5 & \cellcolor{timecol!27}.054 \\
        7 & CosyPose & Mask R-CNN & 2020 & \cellcolor{ccol!100}RGB & \cellcolor{ccol!100}PBR & \cellcolor{avgcol!55.7}55.7 & \cellcolor{timecol!28}.055 \\
        8 & FCOS-CDPN & FCOS & \cellcolor{ccol!100}2022 & \cellcolor{ccol!100}RGB & \cellcolor{ccol!100}PBR & \cellcolor{avgcol!50.7}50.7 & \cellcolor{timecol!24}.047 \\ 
        \bottomrule
    \end{tabularx}
    \caption{\textbf{2D object detection results.}
    The methods are ranked by the $\text{AP}_C$ score defined in Sec.~\ref{sec:2d_tasks}. The last column shows the average image processing time (in seconds).
    }
    \label{tab:det_results}
    \vspace{1.8ex}
\end{table}

\subsection{2D Object Segmentation Results} \label{sec:seg_results}
\setlength{\tabcolsep}{2pt}
\begin{table}[t]
    \footnotesize
    \begin{tabularx}{\columnwidth}{lLllllll} \toprule
        \# & Method & ...based on & Year & Data & ...type & AP$_C$ & Time \\ 
        \toprule
        1 & ZebraPoseSAT & CosyPose+Zebra & \cellcolor{ccol!100}2022 & \cellcolor{ccol!100}RGB & PBR+real & \cellcolor{avgcol!58.7}58.7 & \cellcolor{timecol!40}.080 \\ 
        2 & ZebraPoseSAT & CDPNv2+Zebra & \cellcolor{ccol!100}2022 & \cellcolor{ccol!100}RGB & PBR+real & \cellcolor{avgcol!57.8}57.8 & \cellcolor{timecol!40}.080 \\ 
        3 & ZebraPoseSAT & CosyPose+Zebra & \cellcolor{ccol!100}2022 & \cellcolor{ccol!100}RGB & \cellcolor{ccol!100}PBR & \cellcolor{avgcol!53.8}53.8 & \cellcolor{timecol!40}.080 \\ 
        4 & ZebraPoseSAT & CDPNv2+Zebra & \cellcolor{ccol!100}2022 & \cellcolor{ccol!100}RGB & \cellcolor{ccol!100}PBR & \cellcolor{avgcol!52.3}52.3 & \cellcolor{timecol!40}.080 \\ 
        5 & DLZDet & DLZDet & \cellcolor{ccol!100}2022 & \cellcolor{ccol!100}RGB & PBR+real & \cellcolor{avgcol!49.6}49.6 & - \\ 
        6 & DLZDet & DLZDet & \cellcolor{ccol!100}2022 & \cellcolor{ccol!100}RGB & \cellcolor{ccol!100}PBR & \cellcolor{avgcol!42.9}42.9 & - \\ 
        7 & CosyPose & Mask R-CNN & 2020 & \cellcolor{ccol!100}RGB & PBR+real & \cellcolor{avgcol!40.5}40.5 & \cellcolor{timecol!27}.054 \\ 
        8 & CosyPose & Mask R-CNN & 2020 & \cellcolor{ccol!100}RGB & \cellcolor{ccol!100}PBR & \cellcolor{avgcol!36.2}36.2 & \cellcolor{timecol!28}.055 \\ 
        \bottomrule
    \end{tabularx}
    \caption{\textbf{2D object segmentation results.}
    Details as in Tab.~\ref{tab:det_results}.
    }
    \label{tab:seg_results}
    \vspace{0.5ex}
\end{table}

We see an improvement from 40.5~AP$_C$ achieved by the default masks from Mask R-CNN to 58.7~AP$_C$ achieved by masks from ZebraPoseSAT~\cite{su2022zebrapose} (+18.2 AP$_C$; \#1$-$\#7 in Tab.~\ref{tab:seg_results}).
Interestingly, ZebraPoseSAT predicts the high-quality masks in regions determined by the default detections from Mask R-CNN (\#6 in Tab.~\ref{tab:det_results}) and would likely achieve even higher segmentation accuracy if relying on detections from YOLOX trained for GDRNPP.
As mentioned in Sec.~\ref{sec:6d_results}, most 6D object localization methods evaluated in 2022 start by 2D object detection. Leveraging 2D object segmentation instead could improve results on objects with irregular shapes~\cite{yang2021ishape} which are included, \eg, in the industrial ITODD dataset~\cite{drost2017introducing}.

\section{Awards} \label{sec:awards}

The following BOP Challenge 2022 awards were presented at the 7th Workshop on Recovering 6D Object Pose\footnote{
\texttt{\href{https://cmp.felk.cvut.cz/sixd/workshop_2022/}{cmp.felk.cvut.cz/sixd/workshop\_2022}}} organized at the ECCV 2022 conference. The awards are based on the 6D object localization results in Tab.~\ref{tab:pose_methods_results}, method properties in Tab.~\ref{tab:pose_methods_properties}, the 2D object detection results in Tab.~\ref{tab:det_results}, and the 2D object segmentation results in Tab.~\ref{tab:seg_results}.

The \emph{GDRNPP}~\cite{Wang_2021_GDRN, liu2022gdrnpp_bop} submissions were prepared by Xingyu Liu, Ruida Zhang, Chenyangguang Zhang, Bowen Fu, Jiwen Tang, Xiquan Liang, Jingyi Tang, Xiaotian Cheng, Yukang Zhang, Gu Wang, Xiangyang Ji; \emph{Extended\_FCOS+PFA}~\cite{hu2022perspective} by Yang Hai, Rui Song, Zhiqiang Liu, Jiaojiao Li, Mathieu Salzmann, Pascal Fua, Yinlin Hu; \emph{ZebraPoseSAT}~\cite{su2022zebrapose} by Yongzhi Su, Praveen Nathan, Torben Fetzer, Jason Rambach, Didier Stricker, Mahdi Saleh, Yan Di, Nassir Navab, Benjamin Busam, Federico Tombari, Yongliang Lin, Yu Zhang, \emph{Coupled Iterative Refinement}~\cite{lipson2022coupled} by Lahav Lipson, Zachary Teed, Ankit Goyal, and Jia Deng; and \emph{RCVPose3D}~\cite{wu2022keypoint} by Yangzheng Wu, Alireza Javaheri, Mohsen Zand, Michael Greenspan.

\vspace{0.7em}
\noindent Awards for 6D object localization methods:
\begin{itemize}
    \setlength\itemsep{-0.3em}
    \item \textbf{The Overall Best Method:} \\ \textit{GDRNPP-PBRReal-RGBD-MModel}
    \item \textbf{The Best RGB-Only Method:} \\ \textit{GDRNPP-PBRReal-RGB-MModel}
    \item \textbf{The Best Fast Method (less than 1s per image):} \\ \textit{GDRNPP-PBRReal-RGBD-MModel-Fast}
    \item \textbf{The Best BlenderProc-Trained Method:} \\ \textit{GDRNPP-PBR-RGBD-MModel}
    \item \textbf{The Best Single-Model Method} (trained per dataset)\textbf{:} \\ \textit{Extended\_FCOS+PFA-MixPBR-RGBD}
    \item \textbf{The Best Open-Source Method:} \\ \textit{GDRNPP-PBRReal-RGBD-MModel}
    \item \textbf{The Best Method On Default Detections/Segment.:} \\ \textit{GDRNPP-PBRReal-RGBD-MModel-OfficialDet}
    \item \textbf{The Best Method on T-LESS, ITODD, YCB-V, HB:} \\ \textit{GDRNPP-PBRReal-RGBD-MModel}
    \item \textbf{The Best Method on LM-O:} \\ \textit{Extended\_FCOS+PFA-MixPBR-RGBD}
    \item \textbf{The Best Method on TUD-L:} \\ \textit{Coupled Iterative Refinement (CIR)}
    \item \textbf{The Best Method on IC-BIN:} \\ \textit{RCVPose3D\_SingleModel\_VIVO\_PBR}
\end{itemize}

\noindent Awards for 2D object detection/segmentation methods:
\begin{itemize}
    \setlength\itemsep{-0.3em}
    \item \textbf{The Overall Best Detection Method:} \\ \textit{GDRNPPDet\_PBRReal}
    \item \textbf{The Best BlenderProc-Trained Detection Method:} \\ \textit{GDRNPPDet\_PBR}    

    \item \textbf{The Overall Best Segmentation Method:} \\ \textit{ZebraPoseSAT-EffnetB4 (DefaultDetection)}
    \item \textbf{The Best BlenderProc-Trained Segment. Method:} \\ \textit{ZebraPoseSAT-EffnetB4 (DefaultDet+PBR\_Only)}

    \setlength\itemsep{-0.3em}
\end{itemize}

\section{Conclusions} \label{sec:conclusion}

In the BOP Challenge 2022, we witnessed another breakthrough in the 6D pose estimation accuracy, efficiency and synthetic-to-real transfer.
Methods based on deep neural networks now clearly surpass the traditional methods based on point pair features in both accuracy and speed.
Variations of the winning GDRNPP method~\cite{Wang_2021_GDRN,liu2022gdrnpp_bop} allowed us to analyze the importance of different aspects related to training domains, modalities and run-time efficiency. Besides, we individually measured 2D detection and segmentation performance and could thereby determine sources of gains in the multi-stage pose estimation pipelines. 
Despite the progress, accuracy scores have not been saturated on most BOP datasets and we are already looking forward to insights from the next challenge. The online evaluation system at \texttt{\href{http://bop.felk.cvut.cz/}{bop.felk.cvut.cz}} stays open and raw results of all methods will be made publicly available.

{\small
\bibliographystyle{ieee_fullname}
\bibliography{references}
}

\end{document}